%% file: main.tex
\documentclass[runningheads]{llncs}

\usepackage{eccv}

\usepackage{eccvabbrv}

\usepackage{graphicx}
\usepackage{booktabs}

\usepackage[accsupp]{axessibility}  

\usepackage{colortbl}
\usepackage{makecell}
\usepackage{tikz}
\usepackage{import}

\usepackage{hyperref}

\usepackage{orcidlink}

\begin{document}

\title{Beyond Classification: Task-Dependent Learnability under Privacy-Motivated Image Transformations} 
\titlerunning{Beyond Classification: Task-Dependent Learnability}

\author{
  Leon Ranke\inst{1, 2}\orcidlink{0009-0005-9312-3441} \and
  Wolfgang Hübner\inst{1}\orcidlink{0000-0001-5634-6324} \and
  Ronny Hug\inst{1}\orcidlink{0000-0001-6104-710X} \and
  Michael Arens\inst{1}\orcidlink{0000-0002-7857-0332} \and
  Jürgen Beyerer\inst{1, 2}\orcidlink{0000-0003-3556-7181}
}

\authorrunning{L.~Ranke et al.}

\institute{
  Fraunhofer Institute of Optronics, System Technologies and Image Exploitation (IOSB), Gutleuthausstraße~1, 76275 Ettlingen, Germany \and
  Karlsruhe Institute of Technology (KIT)
  \email{firstname.lastname@iosb.fraunhofer.de}\\
  \url{https://www.iosb.fraunhofer.de}
}

\maketitle

\begin{abstract}
  Privacy-Enhancing Technologies (PETs) in computer vision often rely on noise or image perturbations to protect visual data while securely processing it, creating a trade-off between task performance and protection. This trade-off is commonly evaluated using image classification, which primarily captures semantic separability and remains robust despite significant geometric, spatial layout or local boundary alterations. As a result, it is too simplistic as a proxy for generic vision tasks. Exhaustive downstream-task evaluation, however, is computationally expensive because models must often be trained for each PET transformation and parameter setting. We therefore propose a compute-aware multi-task protocol for evaluating PETs in model training. It combines lightweight proxy tasks that target complementary aspects of visual structure while remaining simple and fast to compute. Across irreversible privacy transformations, key-based block primitives, and learnable image encryption schemes, we demonstrate that PETs with similar classification accuracy can differ substantially on other tasks. The outcomes highlight the need for PET evaluation protocols that move beyond classification-only reporting. Code is available at: \url{https://github.com/LeonRanke/Task-Dependent-Learnability}.
  
\end{abstract}

\section{Introduction}
\label{sec:Introduction}

Privacy-Preserving Machine Learning (PPML), as applied to computer vision tasks, aims to reduce the exposure of sensitive visual content while retaining sufficient information for effective learning. For representation-altering PETs (Privacy-Enhancing Technologies), such as learnable image encryption \cite{Tanaka2018_LE, Madono2020_ELE, Hirose2025_LEViT} and related image-domain transformations \cite{Kurihara2015_EtC, Huang2020_Instahide, Sirichotedumrong2019_PbIE}, this creates a fundamental tension: the transformation should obscure human-interpretable information, while the resulting representation must still support downstream learning.

Existing evaluations of representation-altering PETs commonly report classification accuracy as the main measure of utility \cite{Tanaka2018_LE, Madono2020_ELE, Huang2020_Instahide, Kurihara2015_EtC}. This provides a useful but narrow view: classification primarily tests whether class-discriminative statistics remain separable under the transformed data distribution. It does not necessarily probe whether geometric relations, spatial layout, boundary compatibility, or orientation-dependent structures are preserved \cite{Zamir2018_Taskonomy}. Indeed, models can maintain high classification accuracy even when spatial structure and high-frequency components are substantially degraded \cite{Geirhos2018_Texturebias, Brendel2019_BagNets, Hermann2020_OriginsTexturebias}. Transformed images may therefore support classification while being poorly suited for tasks such as detection, segmentation, tracking, pose estimation, or geometric matching.

A natural solution is to evaluate PETs on a broad suite of downstream tasks, but in practice this is often prohibitively expensive. Because transformations alter low-level statistics and spatial structure, pre-training on natural images cannot be assumed to generalise across parametrisations, including encryption keys, design parameters, and scale. Rigorous evaluation would therefore require training from scratch for each transformation family and parameter setting. This motivates a compute-aware intermediate evaluation stage: before investing in expensive downstream pipelines, lightweight diagnostic proxy tasks (\eg relative angle prediction or jigsaw puzzle-solving) can probe which broad types of visual structure remain learnable.

In this paper, \emph{learnability} is used in an operational, empirical sense. Given a downstream task, a transformation $T_{\lambda, \kappa}$,\footnote{Elements of a transformation family are indexed by the parameter vector $\lambda$; $\kappa$ denotes a key, if applicable.} a model, and a training protocol, learnability is estimated as the task performance achieved by a model trained from scratch on transformed samples $\widetilde{x} = T_{\lambda, \kappa}(x)$. Learnability is thus not an intrinsic property of the transformation alone, instead it characterises what information a particular learning algorithm can access under finite data and computational resources. In this light we distinguish \emph{obfuscation} from \emph{encryption}: the former irreversibly alters an image, whereas the latter is invertible given the key $\kappa$ (security is not part of this definition). Information may be preserved in principle yet remain inaccessible, or difficult to exploit, to a model that receives no key-dependent information.

Our results show that learnability under privacy-motivated image transformations is strongly task-dependent. Transformations that preserve comparatively high classification accuracy can severely impair tasks that rely on geometric consistency and spatial order. Conversely, key-based transformations can introduce deterministic shortcut cues that some tasks exploit, resulting in high performance, while performance on other tasks collapses. By relating these task-specific behaviours to image-domain obfuscation and leakage metrics, we find that no single utility or obfuscation metric fully characterises a transformation. Together, these findings motivate evaluation protocols for privacy-preserving vision that go beyond classification performance, jointly assessing task-dependent learnability, shortcut leakage, and image-domain obfuscation.

In summary, our main contributions are:
\begin{enumerate}
    \item A compute-aware multi-task protocol for evaluating task-dependent learnability under privacy-motivated image transformations.
    \item An analysis of a controlled set of multi-scale image transformations, spanning obfuscation, block-based transforms and Learnable Image Encryption schemes \cite{Tanaka2018_LE, Madono2020_ELE, Kurihara2015_EtC}, showing how scale and key structure affect learnability.
    \item Evidence that classification accuracy, proxy-task utility, and image-domain obfuscation metrics capture distinct and individually incomplete aspects of representation-altering PETs in PPML.
\end{enumerate}

\section{Related Work}
\label{sec:RelatedWork}

Research on PPML for computer vision has primarily focused on secure computation mechanisms \cite{Lee2023_FHEResNet, Juvekar2018_Gazelle, Gilad-Bachrach2016_CryptoNets, Lou2020_Glyph, Jarin2021_Pricure}, privacy-preserving training protocols \cite{Dong2020_DPoptimal, Phuong2023_DPSGD, Zuo2024_FedViTit, Wen2023_FLsurvey}, and data transformations \cite{Tanaka2018_LE, Madono2020_ELE, Kurihara2015_EtC, Huang2020_Instahide, Hirose2025_LEViT}. Comparatively little attention has been devoted to systematic evaluation methodologies that jointly assess image-domain obfuscation and downstream task utility.

\subsection{Utility Evaluation in PPML}
\label{sec:UtilityEval}

The notion of model utility varies across PETs. Here, utility refers to how well a protected training pipeline preserves the predictive performance of an equivalent plaintext pipeline. A useful distinction can be drawn between (i) \textit{functionality-preserving} approaches, which aim to reproduce plaintext computations in protected domains; (ii) \textit{training-oriented} PETs, modifying the optimisation process; and (iii) \textit{representation-altering} PETs, that transform images before training.


Functionality-preserving PETs include Fully Homomorphic Encryption \cite{Gentry2009_FHE}, Secure Multiparty Computation \cite{Goldreich2019_SMPC} and Trusted Execution Environments \cite{Ohrimenko2016_TEE}. These approaches execute computations in protected domains while reproducing the behaviour of the corresponding plaintext computation. Utility degradation therefore arises mainly from practical constraints, including reduced numerical precision, approximated non-linear functions, and restricted model architectures. Image classification on benchmarks such as MNIST and CIFAR-10 remains feasible \cite{Gilad-Bachrach2016_CryptoNets,Juvekar2018_Gazelle, Mohassel2017_Secureml, Zhang2024_Heprune, Zhou2024_SMPCsurvey, Schneider2025_HESecureNet, Shafran2021_Crypto, Kucur2025_Privacy}, although often at substantial computational and communication overhead, which can hinder scalability.

Training-oriented PETs include Differential Privacy (DP) \cite{Dwork2006_DP} and Federated Learning (FL) \cite{McMahan2017_FL}. These approaches retain the original input representation while modifying the optimisation or collaboration process. DP-based training introduces a direct privacy--utility trade-off through mechanisms such as gradient clipping and adding calibrated noise \cite{Bu2020_DP, Phuong2023_DPSGD}, whereas FL distributes optimisation across multiple data holders, where utility can degrade through noisy updates, heterogeneous client data, or partial client participation \cite{Wen2023_FLsurvey, Zuo2024_FedViTit}.

Representation-altering PETs include learnable image encryption \cite{Tanaka2018_LE, Madono2020_ELE, Hirose2025_LEViT} and related image-domain transformations \cite{Kurihara2015_EtC, Sirichotedumrong2019_PbIE, Huang2020_Instahide}. These approaches transform images before training and require models to learn directly from the transformed representation. Unlike functionality-preserving PETs, they do not generally preserve arbitrary plaintext computations. Their utility, therefore, depends on whether the transformation retains sufficient task-relevant information, and it has predominantly been evaluated through downstream classification performance relative to training on plaintext images (see \cref{tab:le_datasets}).

This dominance of classification benchmarks creates an evaluation gap for representation-altering PETs. Classification accuracy indicates whether semantic labels remain predictable, but provides limited evidence about other structural properties of the transformed representation. In particular, it does not directly assess whether global orientation, spatial compatibility, local boundary continuity, or position-dependent cues are preserved \cite{Zamir2018_Taskonomy}. These properties matter for many downstream vision tasks, yet they are rarely examined because training complete downstream models in transformed domains is computationally expensive. We address this gap by introducing diagnostic proxy tasks that probe complementary structural requirements while remaining inexpensive enough to evaluate across multiple parameter and key settings. The tasks are adapted from self-supervised objectives \cite{Gidaris2018_AngPred} but serve here as diagnostics of structural learnability rather than pre-training objectives.

\begin{table}[tb]
  \centering
  \caption{Datasets used to evaluate model utility in representation-altering PETs.}
  \label{tab:le_datasets}
  
  \begin{tabular}{l @{\hspace{1cm}} l}
    \toprule
    \textbf{Representation-altering scheme} 
    & \textbf{Dataset(s) used to evaluate utility} \\
    
    \midrule
    
    Learnable Image Encryption~\cite{Tanaka2018_LE} 
    & CIFAR-10~\cite{Krizhevsky2009_CIFAR} \\

    Pixel-Based Image Encryption~\cite{Sirichotedumrong2019_PbIE}
    & CIFAR-10~\cite{Krizhevsky2009_CIFAR}, STL-10~\cite{coates2011_STL} \\
     
    \makecell[l]{Block-wise Scrambled Image \\ Recognition~\cite{Madono2020_ELE}}
    & CIFAR-10, CIFAR-100~\cite{Krizhevsky2009_CIFAR} \\
     
    InstaHide~\cite{Huang2020_Instahide}
    & \makecell[l]{CIFAR-10, CIFAR-100~\cite{Krizhevsky2009_CIFAR}, \\ MNIST~\cite{LeCun1998_MNIST}, ImageNet~\cite{Deng2009_Imagenet} (subset)} \\
     
    \makecell[l]{Learnable Image Encryption for \\ Vision Transformers~\cite{Hirose2025_LEViT}}
    & CIFAR-10~\cite{Krizhevsky2009_CIFAR}, Tiny-ImageNet~\cite{Ye2015_TinyImageNet}\\
  \bottomrule
  \end{tabular}
\end{table}

\subsection{Image Security Metrics}
\label{sec:Security_Metrics}

In contrast to utility evaluation, the image encryption literature provides extensive statistical security evaluation suites \cite{Mahalakshmi2025_ImgEncTechniques, Saberikamarposhti2024_ImgEncTaxonomy}. Commonly reported metrics quantify image-domain properties associated with leakage or obfuscation, including perceptual leakage between plain and transformed images (\eg, MSE, PSNR, SSIM \cite{Wang2004_SSIM}, LPIPS \cite{Zhang2018_LPIPS}, and MI), encrypted-domain statistics (\eg, entropy, adjacent-pixel correlation, spectral flatness, divergence from uniformity), and differential or key-sensitivity properties. \Cref{tab:SecurityMetrics} summarises these metrics.

We treat these quantities as indicators of image-domain obfuscation and leakage rather than as formal privacy guarantees. They measure perceptual, statistical, or differential properties of transformed images, but, by themselves, they do not establish resistance to de-obfuscation, reconstruction, recognition, attribute inference, or membership inference attacks \cite{McPherson2016_DeObfuscation, Zhu2019_DeepLeakage, Geiping2020_InvertingGradients, Shokri2017_MembershipInferenceAttacks, Melis2019_ExploitingFeatureLeakage}. Nevertheless, they are commonly used in image-encryption evaluations \cite{Mahalakshmi2025_ImgEncTechniques, Saberikamarposhti2024_ImgEncTaxonomy} and provide useful complementary information about how strongly a transformation alters the visible image domain. 

\begin{table}[tb]
  \centering
  \caption{Image-domain obfuscation and leakage indicators for RGB images. Metrics are computed per channel (R, G, B) and averaged. Intensities are in $\{0, \dots, L - 1\}$, with $L = 256$ throughout.}
    \resizebox{0.965\linewidth}{!}{
    \begin{tabular}{rlcc}
        \toprule
        \textbf{Name} & \textbf{Abbreviation} & \textbf{Range} & \textbf{Heuristic target} \\
        \hline
        \rowcolor{gray!25}
        \multicolumn{4}{l}{Perceptual Leakage} \\
        Mean Squared Error  & MSE & $[0, (L-1)^2]$  & $\frac{L^2 - 1}{6}$ \\
        Peak Signal-to-Noise Ratio (dB) & PSNR  & $[0, \infty)$  & $10\log_{10}\!\frac{6(L-1)}{L+1}$ \\
        Structural Similarity Index Measure \cite{Wang2004_SSIM} & SSIM & $[-1, 1]$ & $0$  \\
        Learned Perceptual Image Patch Similarity \cite{Zhang2018_LPIPS} & LPIPS & $[0, \infty)$ & $\uparrow$  \\
        Mutual Information (bits)  & MI & $[0, \log_2(L)]$ & $0$  \\
        \hline
        \rowcolor{gray!25}
        \multicolumn{4}{l}{Encrypted Image Statistics} \\
        Chi-Squared Goodness-of-Fit Test & $\chi^2$ & $[0,\infty)$ & $\le \chi^2_{0.05}(L{-}1)$  \\
        Kullback--Leibler Divergence & $D_{\mathrm{KL}}$ & $[0,\infty)$  & $0$ \\
        Shannon Information Entropy (bits) & IE  & $[0, \log_2(L)]$ & $\log_2(L)$ \\
        2D Information Entropy~\cite{Larkin2016_Delentropy} (bits) & 2D-IE & $[0, \log_2(2L{-}1)]$ & $\log_2(2L{-}1)$\\
        Adjacent-pixel correlation & CC  & $[-1, 1]$ & $0$ \\
        Spectral Flatness Measure  & SFM & $[0, 1]$ & $1$ \\
        \hline
        \rowcolor{gray!25}
        \multicolumn{4}{l}{Differential Analysis} \\
        Number of Pixels Change Rate (w.r.t. $\mathcal{I}$)  & $\text{NPCR}_\mathcal{I}$  & $[0, 1]$ & $1 - \frac{1}{L}$  \\
        Unified Average Changing Intensity (w.r.t. $\mathcal{I}$) & $\text{UACI}_\mathcal{I}$  & $[0, 1]$ & $\frac{L + 1}{3L}$ \\
        \hline
        \rowcolor{gray!25}
        \multicolumn{4}{l}{Key Analysis} \\
        Key-space size & $|\mathcal{K}|$  & $[1, \infty)$ & $\ge 2^{128}$  \\
        Number of Pixels Change Rate (w.r.t. $\kappa$)  & $\text{NPCR}_{\kappa}$  & $[0, 1]$ & $1 - \frac{1}{L}$ \\
        Unified Average Changing Intensity (w.r.t. $\kappa$) & $\text{UACI}_{\kappa}$ & $[0, 1]$ & $\frac{L + 1}{3L}$ \\
        \bottomrule
    \end{tabular}}
    \label{tab:SecurityMetrics}
\end{table}

\section{Methodology}
\label{sec:Methodology}

The objective of this paper is to characterise how image-domain transformations affect the learnability of different visual structures. To this end, transformations are viewed as operators that remove, rearrange, or statistically alter components of the visual signal. Downstream proxy tasks then act as diagnostic probes that test whether these components remain accessible to the learning algorithm.

Transformations are defined as parametrisable operators $T_{\lambda, \kappa}: \mathcal{X} \rightarrow \widetilde{\mathcal{X}}$ acting on the plain image space $\mathcal{X} \subset \mathbb{R}^{H\!\times\!W\!\times\!3}$, where $\lambda$ denotes the scale parameter and $\kappa \in \mathcal{K}$ a secret key. For non-keyed obfuscations, $\kappa = \varnothing$. Each transformation induces a modified data distribution with altered  spatial, geometric, perceptual, or statistical properties.

\subsection{Transformation Operators}
\label{sec:Transformations}

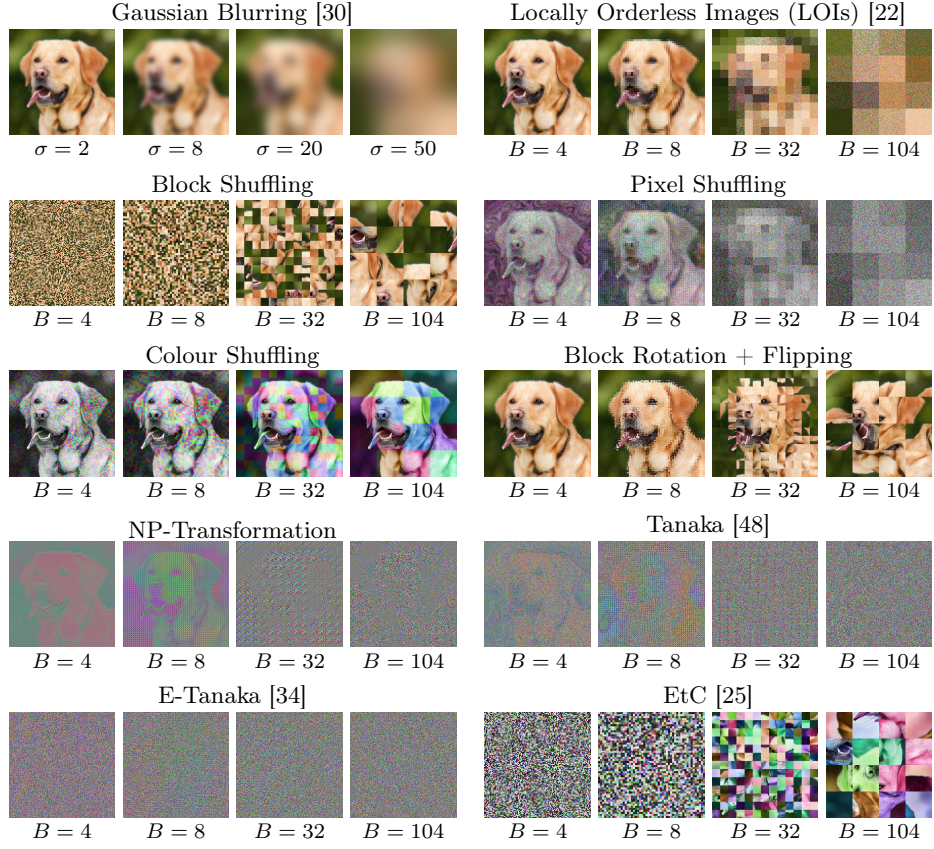
\begin{figure}[tb]
    \centering
    \makebox[\textwidth]{\import{img/}{Examples.pgf}}
    \caption{Qualitative examples of transformation families across increasing scale.}
    \label{fig:Example_Transformations}
\end{figure}

We evaluate controlled irreversible transformations, key-based block primitives, and learnable image encryption schemes \cite{Tanaka2018_LE,Madono2020_ELE,Kurihara2015_EtC}. The first group acts as a diagnostic baseline that removes specific visual structures in a controlled manner, while the remaining two groups represent transformations used in learnable image encryption. Together, they enable a systematic analysis of which structural components remain learnable under different forms of image-domain alteration. \Cref{fig:Example_Transformations} provides qualitative examples of the considered transforms.

\textbf{Gaussian Blurring} acts as a low-pass filter, removing high-frequency components while preserving spatial topology and global intensity relationships. For scale parameter $\sigma$, images are obtained via separable convolution with a Gaussian kernel of standard deviation $\sigma$ and kernel size $k = 2 \lfloor 3 \sigma \rfloor + 1$ \cite{Lindeberg1994_ScaleSpace}. Increasing $\sigma$ progressively suppresses fine-scale details, isolating the contribution of spatial-frequency bands to downstream task performance.

\textbf{Locally Orderless Images (LOIs)} discard spatial ordering within non-overlapping $B \times B$ windows while approximately preserving marginal intensity distributions inside those windows \cite{Koenderink1999_LOI}. As $B$ increases, local correlations are removed at progressively larger scales, while global intensity statistics are retained. LOIs, therefore, provide a controlled mechanism for analysing the impact of spatial order at different scales on learnability. 

\textbf{Block-Based Transformations} are invertible operations used in the considered learnable image encryption schemes \cite{Tanaka2018_LE, Madono2020_ELE, Kurihara2015_EtC}. Images are partitioned into non-overlapping $B \times B$ blocks, followed by deterministic, key-based operations applied within or across blocks. Depending on the primitive, these operations either preserve local block content while disrupting global arrangement, or disrupt local adjacency while preserving block-level placement. The analysed primitives are: permutation of spatial block positions (\emph{Block Shuffling}); block rotations by multiples of $90^\circ$ combined with horizontal and vertical flips (\emph{Block Rotation + Flipping}); permutation of pixel positions within blocks (\emph{Pixel Shuffling}); permutation of the RGB channels per block (\emph{Colour Shuffling}); and pixel-wise intensity inversion (\emph{Negative-Positive (NP) Transformation}).

\textbf{Learnable Image Encryption Schemes} combine multiple block-based primitives into structured pipelines. These schemes are invertible given the secret key $\kappa$, but can induce transformed-domain regularities that are difficult to exploit without access to the secret key. We analyse: \emph{Tanaka} \cite{Tanaka2018_LE} (Pixel Shuffling and NP Transformation); \emph{Extended Tanaka (E-Tanaka)} \cite{Madono2020_ELE} (Pixel Shuffling, NP Transformation, Block Shuffling); and the \emph{Encryption then Compression (EtC)} \cite{Kurihara2015_EtC} scheme (Block Rotation + Flipping, Colour Shuffling, Block Shuffling).

Gaussian Blurring uses $\sigma$ as the scale parameter $\lambda$, whereas all other transformations use the block size $B$. The distinction between obfuscations and keyed transformations matters here: Gaussian blur and LOIs remove information from the image even when their parameters are known, whereas block-based transformations and learnable image encryption schemes preserve information. A learner receiving only transformed images may nonetheless be unable to exploit this information effectively, because the transformation can misalign with the model's inductive biases. Some PETs address this restriction by implicitly providing the model with key-dependent information (\eg, E-Tanaka \cite{Madono2020_ELE} provides the model with information related to the block permutation). Moreover, when a fixed key is used across training and testing, deterministic key-dependent regularities may become learnable and act as shortcut cues.\footnote{See \cref{sec:JigsawResults} for empirical evidence of key-induced shortcut cues.} Our protocol therefore measures fixed-key empirical learnability under a specified training setup, rather than intrinsic information preservation or formal privacy.

\subsection{Evaluation Tasks}
\label{sec:Tasks}

The goal is not to approximate the performance of every possible downstream task. Instead, a tractable protocol is introduced that can be repeated across many transformations, scales, and keys to identify which broad types of visual structure remain learnable. The selected tasks act as diagnostic probes: they are inexpensive relative to full detection or segmentation pipelines and require different structural properties of the transformed image distribution. Rather than serving as independent benchmarks, they function as a combined test suite for structural sufficiency under transformation.

\textbf{Image Classification} predicts a discrete class label $y \in \{0, \dots, N-1\}$ given transformed images $\widetilde{x} = T_{\lambda, \kappa}(x)$. We use a ResNet-34 \cite{He2016_ResNet} with an $N$-class classification head. Since classification provides a sparse supervision signal, a wide range of input variations can be tolerated as long as class-discriminative statistics remain separable. Classification is included because it is the dominant task for evaluating the utility of representation-altering PETs.\footnote{See \cref{sec:UtilityEval} for a discussion of utility evaluation in the PET literature.}

\textbf{Angle Prediction} \cite{Gidaris2018_AngPred} estimates the relative rotation $\Delta\alpha = \alpha_2 - \alpha_1$ between two independently rotated and transformed views of the same image:
\begin{equation}
    \widetilde{x}_1 = T_{\lambda, \kappa}(R_{\alpha_1} \cdot x), \quad \widetilde{x}_2 = T_{\lambda, \kappa}(R_{\alpha_2} \cdot x).
\end{equation}
The angles $\alpha_1, \alpha_2$ are sampled randomly from a uniform distribution, with transformations applied after rotation. A CNN processes the two transformed images concatenated along the channel dimension. A regression head predicts $\left(\sin(\Delta\alpha), \cos(\Delta\alpha)\right)$ to avoid discontinuities in angular regression. This task differs from classification in two ways. First, it requires consistent extraction of orientation-dependent structure from two transformed views of the same image. Second, the target is continuous, making the task more sensitive to gradual degradation of geometric information. 

\textbf{Jigsaw Puzzle Solving} \cite{Noroozi2016_JPS} probes spatial compatibility and local-to-global structural consistency. Given an image split into a regular $k \times k$ grid, the pieces are permuted according to an unknown permutation $\pi$. Following the transformer based JPDVT~\cite{Liu2024_JPDVT}, the model recovers $\pi$ by assigning shuffled pieces to their original grid positions. Varying $k$ changes the spatial scale of the task: larger values reduce the context available per piece and increase the solution space.

Jigsaw puzzle solving tests whether spatial relationships remain learnable after transformation. Downstream tasks such as detection, segmentation, pose estimation, and tracking depend not only on semantic separability but also on spatial layout, boundary compatibility, and consistent relationships between neighbouring regions. Jigsaw solving acts as a tractable intermediate probe for spatially exploitable information. Importantly, high jigsaw accuracy does not necessarily imply preservation of natural spatial structure. Under fixed-key transformations, performance may also reflect deterministic cues induced by the transformations, such as position-dependent block patterns or key-specific regularities. \Cref{tab:task_Overview} summarises the proxy tasks investigated.  

\begin{table}[tb]
  \centering
  \caption{Overview of investigated proxy tasks, including approximate training time.}
  \label{tab:task_Overview}
  \small
  
  \begin{tabular}{llc}
    \textbf{Task} & \textbf{Probes} &  \textbf{Approx. Time}                                \\
    \midrule
    Classification          & semantic separability                      & $\approx 1.4$\,h \\
    Angle Prediction        & global orientation / geometric consistency & $\approx 0.8$\,h \\
    Jigsaw Puzzle Solving   & spatial compatibility / local structure    & $\approx 9.4$\,h \\
  \end{tabular}
\end{table}

\section{Experimental Setup}
\label{sec:ExperimentalSetup}

All models are trained \emph{from scratch} exclusively on transformed data $\widetilde{\mathcal{X}}$. They do not receive paired plaintext images, secret keys, or transformation-specific adaptation modules. This enables comparing transformations within a common learning setup and identifying which types of structure remain accessible to standard learners under limited computational resources and data volumes. 

A predefined set of scale parameters $\lambda$ is applied consistently during both training and testing across all transformation families. For key-based transformations, a fixed-key setting is evaluated: one secret key $\kappa$ is sampled per configuration and used for both training and testing. This aligns with common learnable image encryption evaluations, in which models are trained directly in a static transformed domain. However, it also means that models may exploit deterministic key-dependent regularities. Parameter choices for each experiment are reported in the supplementary material. Training is performed for a predefined maximum number of epochs with early stopping, providing an upper bound on optimisation cost while mitigating overfitting.

For classification, we use the Animal Species Classification dataset \cite{Aashman2023_Animals}, which has $N = 15$ classes, with images resized to $416 \times 416$. The dataset provides sufficient spatial resolution to meaningfully analyse scale-dependent transformations while remaining computationally tractable. Performance is evaluated via top-1 accuracy with a chance level of $1/N$.

Angle prediction is evaluated on the same dataset. Images are rotated and then cropped to exclude hard image edges that leak rotational information. Performance is reported as the mean angular error in degrees. Since the angular error is measured as the shortest circular distance between the predicted and ground-truth relative angle, errors lie in $[0^\circ, 180^\circ]$. For uniformly distributed rotations, the expected error of a random or constant predictor is therefore $90^\circ$.

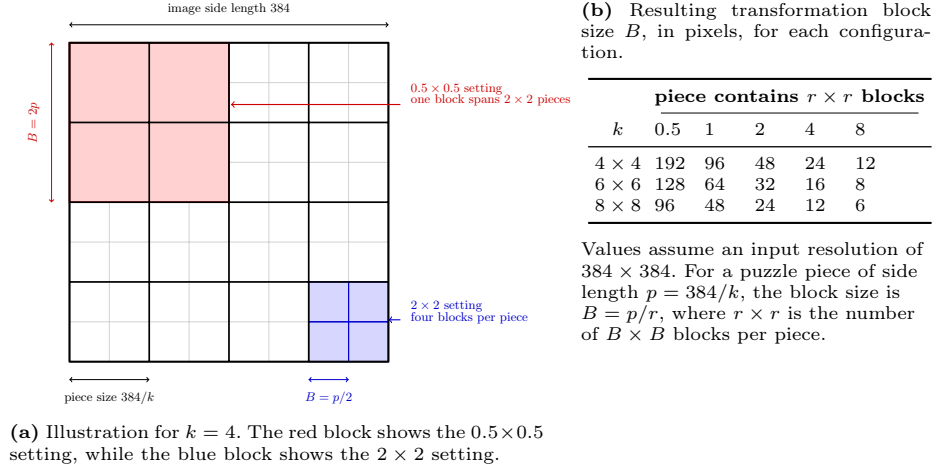
\begin{figure}[t]
    \centering

    \begin{minipage}[t]{0.58\linewidth}
        \vspace{0pt}
        \centering

        \resizebox{\linewidth}{!}{
        \begin{tikzpicture}

            \path[use as bounding box] (-1.4,-1.35) rectangle (11.8,9.25);

            \draw[step=1, very thin, gray!35] (0,0) grid (8,8);

            \fill[red!18] (0,4) rectangle (4,8);
            \draw[red!80!black, very thick] (0,4) rectangle (4,8);

            \fill[blue!16] (6,0) rectangle (8,2);
            \draw[blue!80!black, thick] (6,0) rectangle (7,1);
            \draw[blue!80!black, thick] (7,0) rectangle (8,1);
            \draw[blue!80!black, thick] (6,1) rectangle (7,2);
            \draw[blue!80!black, thick] (7,1) rectangle (8,2);

            \draw[step=2, black, very thick] (0,0) grid (8,8);

            \draw[black, line width=1.2pt] (0,0) rectangle (8,8);

            \draw[<->] (0,8.45) -- (8,8.45);
            \node at (4,8.85) {\small image side length $384$};

            \draw[<->] (0,-0.45) -- (2,-0.45);
            \node at (1,-0.9) {\small piece size $384 / k$};

            \draw[<->, red!80!black] (-0.45,4) -- (-0.45,8);
            \node[red!80!black, rotate=90] at (-0.9,6) {\small $B = 2p$};

            \draw[<->, blue!80!black] (6,-0.45) -- (7,-0.45);
            \node[blue!80!black] at (6.5,-0.9) {\small $B = p/2$};

            \node[
                red!80!black,
                align=left,
                anchor=west
            ] at (8.45,6.7) {
                \small $0.5 \times 0.5$ setting\\[-1mm]
                \small one block spans $2 \times 2$ pieces
            };

            \draw[->, red!80!black, thick] (8.3,6.45) -- (4.05,6.45);

            \node[
                blue!80!black,
                align=left,
                anchor=west
            ] at (8.45,1.25) {
                \small $2 \times 2$ setting\\[-1mm]
                \small four blocks per piece
            };

            \draw[->, blue!80!black, thick] (8.3,1.05) -- (8.0,1.05);

        \end{tikzpicture}
        }
        \subcaption{Illustration for $k = 4$. The red block shows the $0.5 \times 0.5$ setting, while the blue block shows the $2 \times 2$ setting.}
        \label{fig:block_size_illustration}
    \end{minipage}
    \hfill
    \begin{minipage}[t]{0.38\linewidth}
        \vspace{0pt}
        \centering
        \scriptsize

        \subcaption{Resulting transformation block size $B$, in pixels, for each configuration.}
        \label{fig:block_size_table}

        \vspace{0.5em}

        \setlength{\tabcolsep}{2.5pt}
        \resizebox{\linewidth}{!}{
        \begin{tabular}{cp{0.5cm}p{0.5cm}p{0.5cm}p{0.5cm}p{0.5cm}}
            \toprule
            & \multicolumn{5}{c}{\textbf{piece contains $r \times r$ blocks}} \\
            \cmidrule(lr){2-6}
            \textbf{$k$} & $0.5$ & $1$  & $2$  & $4$  & $8$ \\
            \midrule
            $4 \times 4$ & $192$ & $96$ & $48$ & $24$ & $12$ \\
            $6 \times 6$ & $128$ & $64$ & $32$ & $16$ & $8$  \\
            $8 \times 8$ & $96$  & $48$ & $24$ & $12$ & $6$  \\
            \bottomrule
        \end{tabular}}

        \vspace{0.75em}

        \raggedright
        \scriptsize
        Values assume an input resolution of $384 \times 384$. For a puzzle piece of side length $p = 384 / k$, the block size is $B = p / r$, where $r \times r$ is the number of $B \times B$ blocks per piece.
    \end{minipage}

    \caption{Relationship between the puzzle size $k$ and the transformation block size $B$.}
    \label{fig:jigsaw_block_sizes}
\end{figure}

In the puzzle-solving experiments, images from the Animal Species Classification dataset are resized to $384 \times 384$ pixels and divided into a regular $k \times k$ grid, where $k \in \{4, 6, 8\}$ denotes the puzzle size. For each puzzle size, the transformation block size $B$ is chosen relative to the side length $p = 384 / k$ of one puzzle piece. We consider settings in which each puzzle piece contains $r \times r$ blocks, with $r \in \{0.5, 1, 2, 4, 8\}$. The considered configurations and resulting values are shown in \cref{fig:jigsaw_block_sizes}. This construction controls the relationship between the transformation and puzzle scales, avoiding misalignment between transformation blocks and piece boundaries. Performance is measured as the percentage of correctly placed puzzle pieces; the chance level of a $k \times k$ puzzle is $1 / k^2$.

\section{Results \& Discussion}
\label{sec:Results}

This section analyses how the considered transformations affect task-dependent learnability. We first report classification performance, which aligns with the dominant utility view in prior work, and then compare it with relative-angle prediction and jigsaw puzzle solving. Finally, we relate task performance to image-domain obfuscation metrics.

\subsection{Image Classification}
\label{sec:ClsResults}

\begin{figure}[tb]
    \centering
    \includegraphics[width=\textwidth]{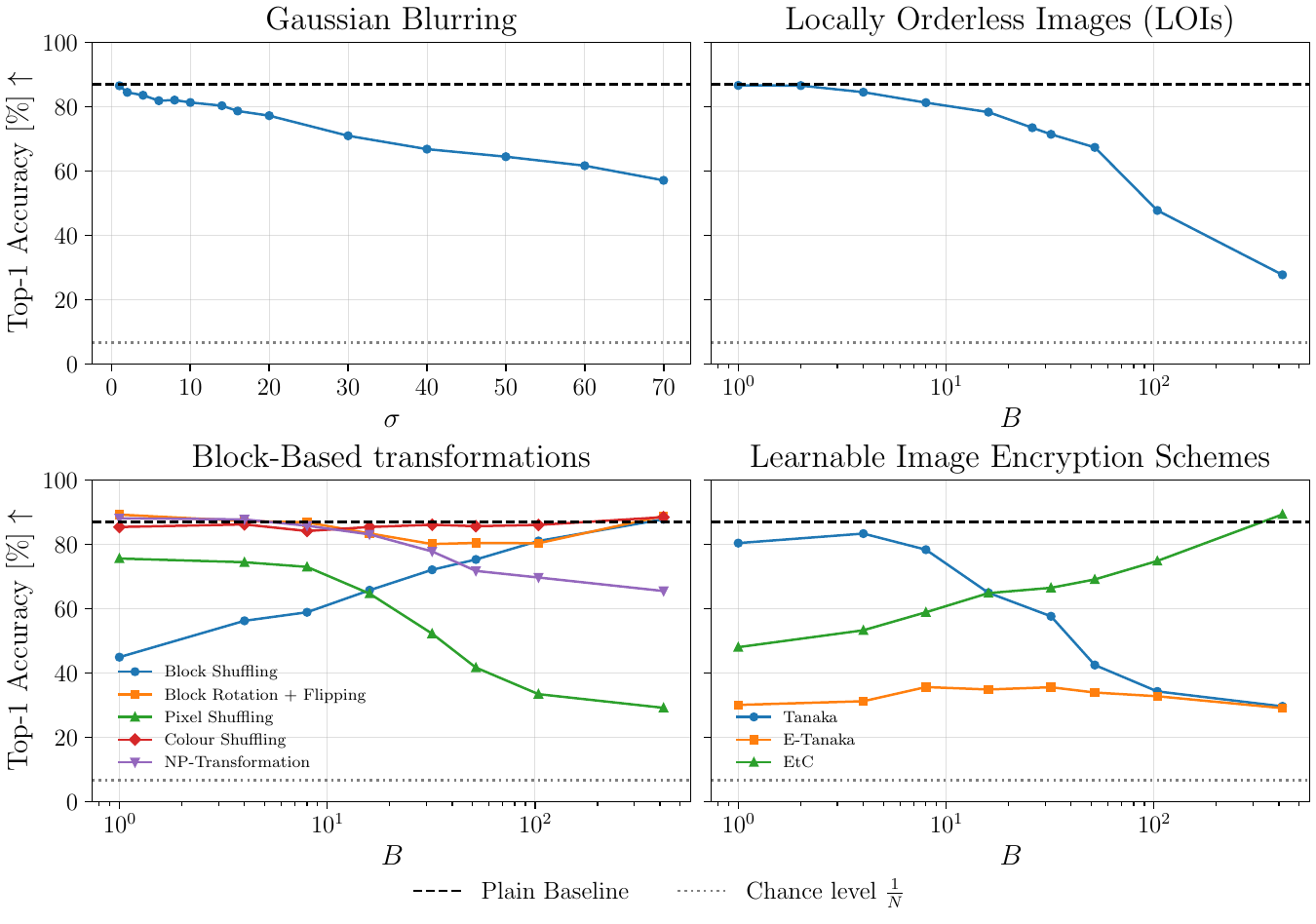}
    \caption{Classification accuracy against transformation strength. Accuracy remains above chance for all transformations, indicating that semantic separability is preserved even under substantial image-domain alteration.}
    \label{fig:ClsResults}
\end{figure}

\Cref{fig:ClsResults} shows the top-1 classification accuracies as $\lambda$ increases. The plain baseline reaches $87.00\%$, while chance performance is $6.67\%$. Classification remains well above chance for all transformations, even under substantial image-domain alteration. Gaussian blur degrades accuracy gradually, but still achieves $57.10\%$ at $\sigma = 70$. LOIs show a stronger scale effect, decreasing from near-baseline performance at small block sizes to $27.75\%$ when correlations are removed globally.

Block-based transformations differ according to which structures they disrupt. Colour shuffling, block rotation/flipping, and the NP-transformation preserve high classification accuracy across most scales, indicating that class discriminative information remains accessible despite substantial visual changes. In contrast, pixel shuffling increasingly disrupts local adjacency and reduces accuracy at larger block sizes. Block shuffling shows the opposite trend: small blocks strongly disrupt the global arrangement, whereas larger blocks preserve more of the layout and therefore recover classification performance.

The composed encryption schemes inherit these behaviours. Tanaka degrades with increasing pixel-shuffling scale; EtC improves as larger shuffled blocks preserve more global structure, and E-Tanaka remains comparatively low across scales because it combines local and global disruption. Overall, classification captures only semantic separability: high accuracy indicates that semantic labels remain predictable, but it does not imply that geometric or spatial structures required by other tasks are preserved. This does not mean classification is always the more permissive probe. As \cref{sec:AngleResults} shows, for some transformations (\eg, Gaussian Blurring), classification degrades faster than angle prediction, so the achieved task utility depends on which structures a transformation removes and which are required for solving the task. Any single task gives an incomplete and potentially misleading picture of utility.

\subsection{Angle Prediction}
\label{sec:AngleResults}

\begin{figure}[tb]
    \centering
    \includegraphics[width=\textwidth]{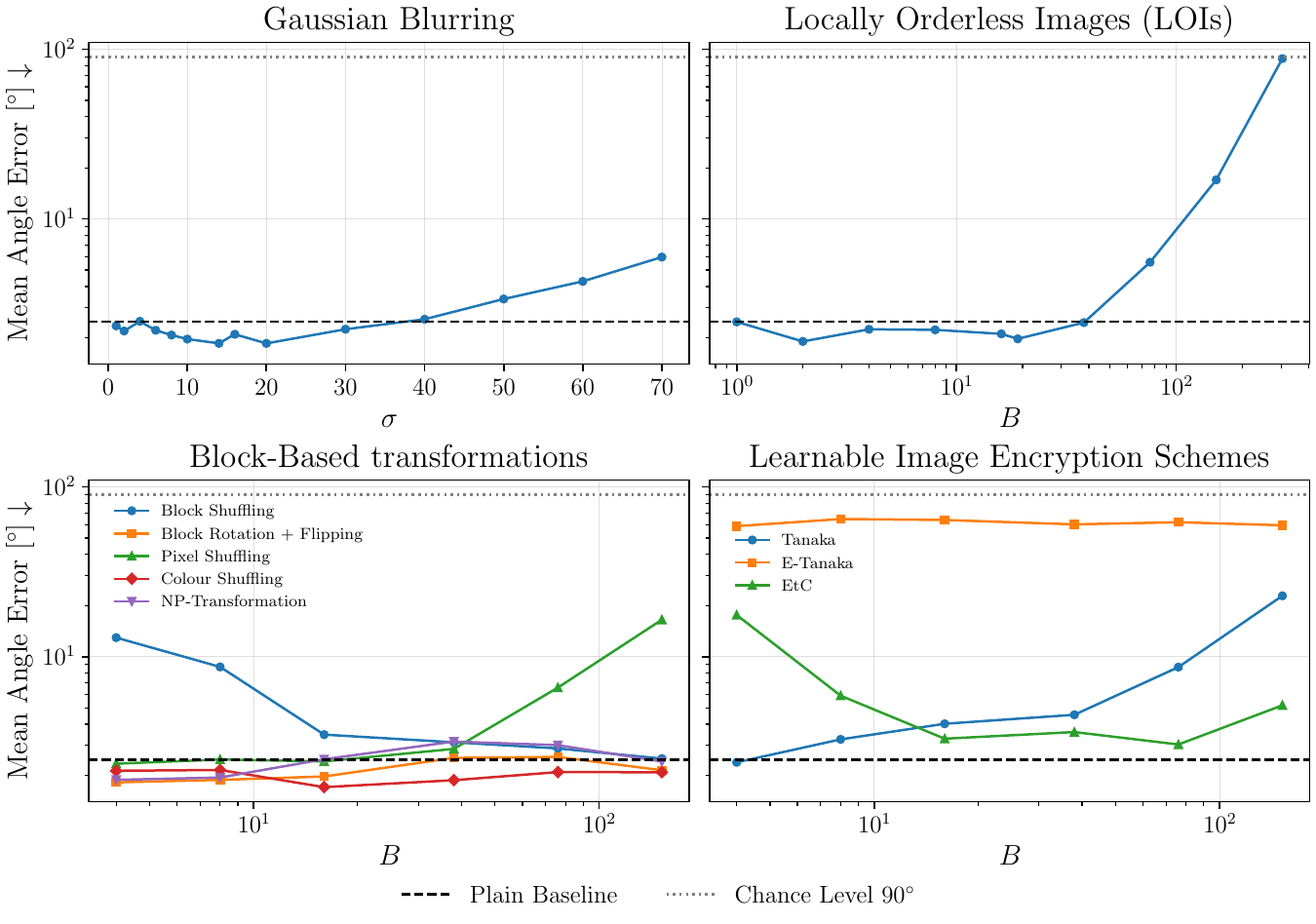}
    \caption{Mean angular error against transform strength (lower is better). Prediction remains robust under transformations that preserve coarse global structure, such as moderate blur, but degrades strongly when spatial order is removed at larger scales or when local and global disruptions are combined.}
    \label{fig:AngResults}
\end{figure}

\Cref{fig:AngResults} reports the mean angular error, with a chance performance of $90^\circ$. The plain baseline achieves $2.48^\circ$. Unlike classification, angle prediction directly tests whether a coherent global reference frame remains accessible.

Gaussian blur has a limited effect except for large scales, suggesting that relative orientation can be inferred from coarse global structure and does not require fine, high-frequency detail. LOIs behave differently: performance remains close to the baseline for small blocks but degrades rapidly once correlations at relevant scales are removed, approaching chance in global settings.  

For key-based primitives, colour shuffling, NP transformation, and block rotation/flipping have little effect on angle prediction, while pixel shuffling degrades at larger block sizes. Block shuffling again shows a scale-dependent inverse trend: small shuffled blocks disrupt global orientation cues, whereas larger blocks preserve enough layout for accurate prediction. Among composed schemes, E-Tanaka substantially impairs angle prediction across all scales, while Tanaka and EtC depend strongly on block size. These results demonstrate that transformations with similar classification accuracy can differ substantially in the geometric information they preserve.

\subsection{Jigsaw Puzzle Solving}
\label{sec:JigsawResults}
\begin{figure}[tb]
    \centering
    \includegraphics[width=\textwidth]{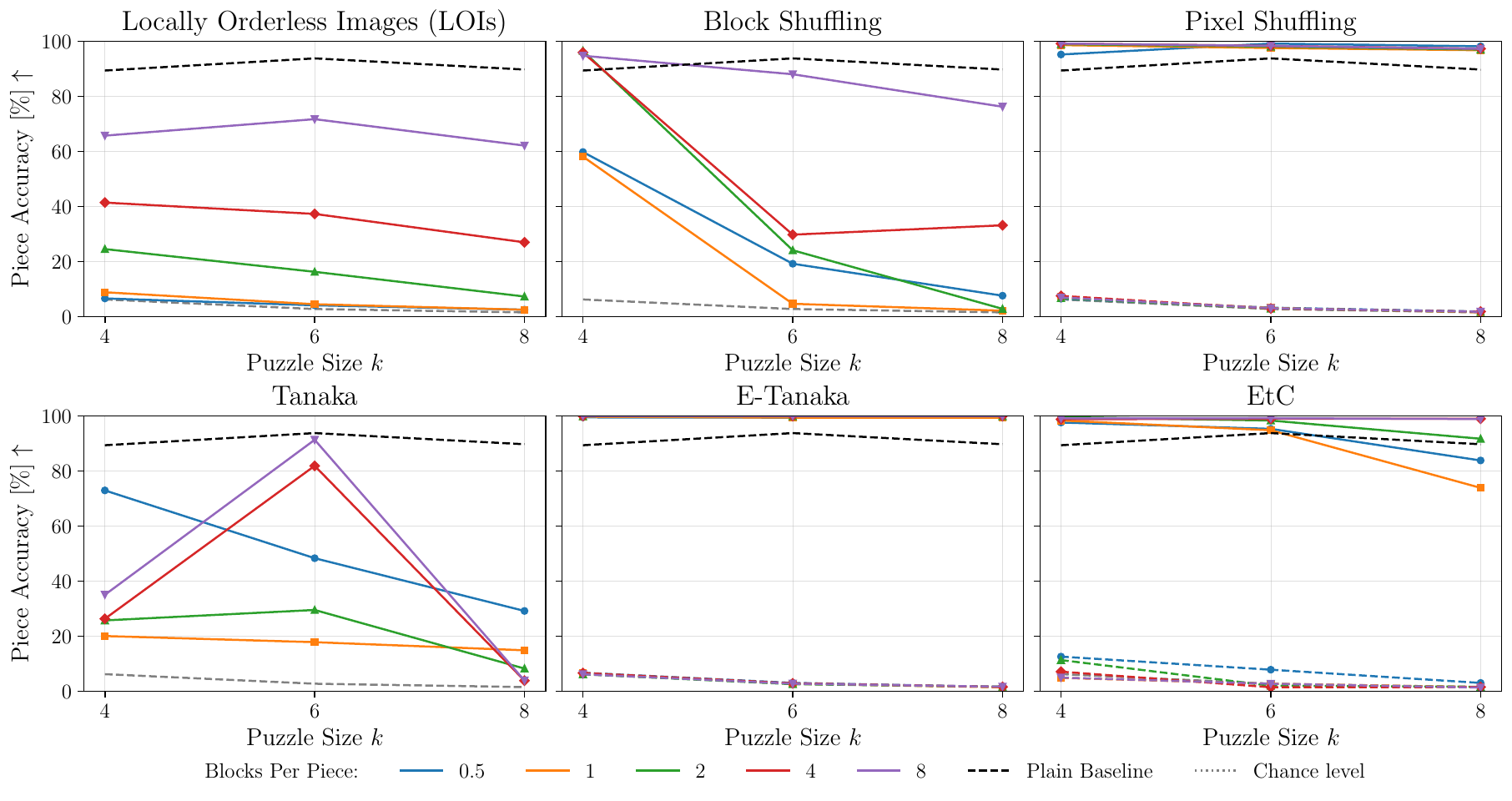}
    \caption{Jigsaw puzzle-solving accuracy for puzzles of different sizes; dashed coloured curves indicate evaluation with a different key from the training key. Puzzle solving exposes spatial placement cues that classification does not capture. LOI and some block-shuffling settings degrade strongly when spatial order is removed at puzzle-relevant scales, whereas several key-based transformations remain highly solvable.}
    \label{fig:JpsResults}
\end{figure}

\Cref{fig:JpsResults} reports piece accuracy for $k \in \{4,6,8\}$ puzzles. The plain baseline achieves high piece accuracy across all puzzle sizes, with chance performance decreasing from $6.25\%$ for $4\times4$ puzzles to $1.56\%$ for $8\times8$ puzzles.

LOI strongly reduces jigsaw performance when spatial order is removed at puzzle-relevant scales. When transformation blocks are as large as, or larger than, puzzle pieces, performance approaches chance. As more transformation blocks fit inside each puzzle piece, local spatial information becomes available, and performance improves. This confirms that the task is sensitive to boundary compatibility and local-to-global spatial consistency.

The key-based transformations reveal a different effect. Pixel shuffling and several composed schemes remain solvable in the fixed-key setting, even when they disrupt global image structure. This does not indicate preservation of natural spatial compatibility. Instead, the puzzle solver can exploit deterministic transformation-induced regularities, such as key-dependent block or position cues that are consistent across training and testing. Evaluating these settings with a key different from the one used in training causes performance to collapse toward chance (see dashed lines in \cref{fig:JpsResults}). Comparing LOI and Pixel Shuffling confirms that the high performance of the latter is due to key-induced regularities. LOI can be viewed as a variant of Pixel Shuffling, with different keys for each block. Attacks on block-scrambling schemes (such as EtC \cite{Kurihara2015_EtC}) use puzzle solvers to reconstruct the plain image \cite{Chuman2023_JigsawAttack}. The high task utility can therefore also be interpreted as a leakage signal. Block shuffling further shows that performance depends on the alignment between transformation scale and puzzle scale: some settings preserve or expose useful placement cues, while others remove them.

Jigsaw solving provides a complementary diagnostic view. Low performance indicates that spatial information required by the puzzle solver is no longer readily accessible. High performance may reflect either the preservation of natural spatial structure or shortcut cues introduced by transformations. This distinction highlights the importance of interpreting tasks jointly rather than in isolation.

\subsection{Obfuscation-Utility Trade-off}

\begin{figure}[tb]
    \centering
    \includegraphics[width=\textwidth]{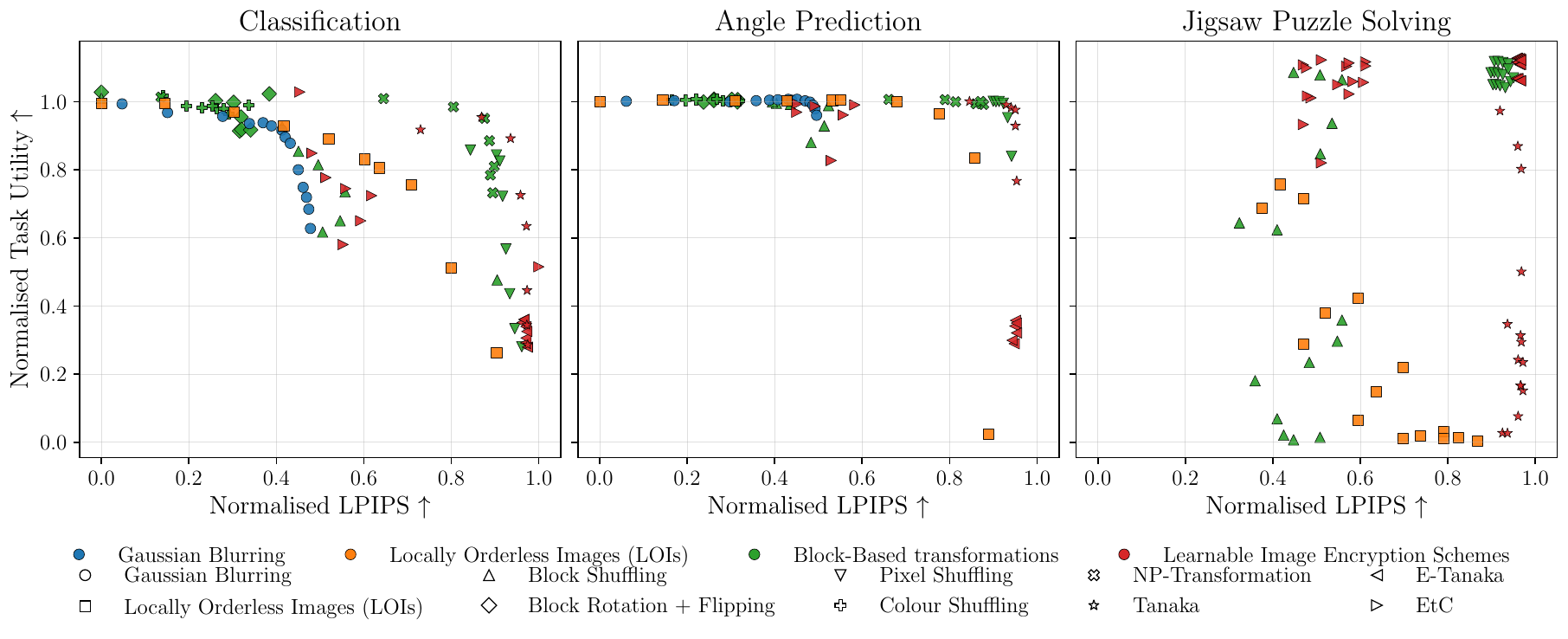}
    \caption{Obfuscation-utility trade-off. Normalised task utility against normalised LPIPS. For angle prediction, the utility direction is inverted, since lower errors are better.}
    \label{fig:UtilityVsSecurity}
\end{figure}

Task utility alone is only one aspect of PET evaluation. A transformation that preserves high task performance while remaining perceptually close to the original provides limited obfuscation; one that strongly alters the image but destroys downstream learnability is equally impractical. Evaluation should therefore jointly consider task utility and image-domain leakage indicators. \Cref{fig:UtilityVsSecurity} relates normalised task utility to an LPIPS-based obfuscation score \cite{Zhang2018_LPIPS}, where higher utility indicates better learnability relative to the plain baseline and higher LPIPS indicates lower perceptual similarity to the original. The metrics in \cref{tab:SecurityMetrics} constitute the broader evaluation framework; LPIPS is used here as a representative perceptual indicator. The full PET evaluation should consider additional metrics (see the supplementary material for the remaining metrics in \cref{tab:SecurityMetrics}). 

The trade-off varies substantially across tasks and transformations. Some transformations retain high classification accuracy while offering limited perceptual obfuscation. Others increase perceptual distance but degrade geometric or spatial learnability. Fixed-key transformations add a further complication, achieving high perceptual obfuscation while still exposing deterministic cues that can be exploited by puzzle-solving. No single point on the obfuscation axis, therefore, predicts utility across tasks: the same LPIPS level corresponds to very different learnability depending on which structures a task requires.

\section{Conclusion}
\label{sec:Conclusion}

We studied learnability under image obfuscation as a task-dependent property. Results show that classification accuracy, the dominant utility measure in PET evaluation, provides only a partial view. Transformations that preserve semantic separability can differ substantially in how well they preserve geometric consistency and spatial compatibility. The proposed protocol combines classification, relative-angle prediction, and jigsaw puzzle solving as lightweight diagnostic probes for complementary structures. Further, we show that fixed-key transformations can introduce deterministic shortcut cues that some tasks exploit. This highlights the need to interpret proxy-task performance jointly rather than in isolation. Future evaluations of privacy-preserving vision methods can benefit from moving beyond classification-only reporting to jointly consider task-dependent learnability, shortcut leakage, and image-domain obfuscation. We establish that the three tasks capture distinct structural properties, but not whether they predict full downstream tasks. Training downstream models on a subset of configurations is the natural next step and would confirm the proxies' practical value.



\bibliographystyle{splncs04}
\bibliography{main}

\clearpage

\section*{Supplementary Material}

\section*{Hyperparameter Settings}

\Cref{tab:hyperparams_exp1,tab:hyperparams_exp2,tab:hyperparams_exp3} list the hyperparameters used for classification, relative angle prediction, and jigsaw puzzle solving, respectively. Unless stated otherwise, PyTorch defaults are used. Block sizes $B$ are chosen as divisors of the respective input size, so that blocks tile each image exactly. All experiments use a single random seed per configuration.
 
\begin{table}[tb]
  \centering
  \caption{Hyperparameters for Animal Species Classification.}
  \label{tab:hyperparams_exp1}
  \small
 
  \begin{tabular}{ll}
    \toprule
    \textbf{Hyperparameter} & \textbf{Value} \\
    \midrule
    Dataset & Animal Species Classification dataset \cite{Aashman2023_Animals} \\
    Number of classes & $15$ \\
    Data augmentation & None \\
    Model architecture & ResNet-34~\cite{He2016_ResNet} \\
    Initial weights & Random (trained from scratch) \\
    Input resolution & $3 \times 416 \times 416$ \\
    Loss function & Cross-entropy \\
    Optimizer & Adam~\cite{Kingma2015_Adam} \\
    Learning rate & $10^{-4}$ \\
    Batch size & $64$ \\
    Max.\ epochs & $100$ \\
    Early stopping & patience 5 on validation accuracy\\
    Learning-rate schedule & Multistep at epochs $[20, 50]$ with $\gamma = 0.1$ \\
    Checkpoint selection & Best validation accuracy \\
    Evaluation metric & Top-1 classification accuracy \\
    Random seed & $73$ \\
    Framework & PyTorch 2.11.0+cu126 \\
    Gaussian blur $\sigma$ & $\{1, 2, 4, 6, 8, 10, 14, 16, 20, 30, 40, 50, 60, 70\}$ \\
    Block size $B$  & $\{1, 2, 8, 16, 26, 32, 104, 208, 416\}$ \\
    Hardware & $1\times$ NVIDIA GeForce RTX 4090 \\
    \bottomrule
  \end{tabular}
\end{table}
 
\begin{table}[tb]
  \centering
  \caption{Hyperparameters for Relative Angle Prediction.}
  \label{tab:hyperparams_exp2}
  \small
 
  \begin{tabular}{ll}
    \toprule
    \textbf{Hyperparameter} & \textbf{Value} \\
    \midrule
    Dataset & Animal Species Classification dataset \cite{Aashman2023_Animals} \\
    Angle sampling & $\alpha_1, \alpha_2 \sim \mathcal{U}[-180^\circ, 180^\circ]$ \\
    Data augmentation & None \\
    Model architecture & CNN ($2 \times$ conv, max-pool, $4 \times$ linear) \\
    Initial weights & Random (trained from scratch) \\
    Input resolution & $3 \times 304 \times 304$ \\
    Loss function & Cosine similarity \\
    Optimizer & Adam \cite{Kingma2015_Adam} \\
    Learning rate & $10^{-4}$ \\
    Batch size & $32$ \\
    Max.\ epochs & $50$ \\
    Early stopping & patience 5 on validation loss \\
    Learning-rate schedule & Reduce on plateau, factor $0.5$, patience $3$ \\
    Checkpoint selection & Lowest validation loss \\
    Evaluation metric & Mean angular error in degrees \\
    Random seed & $42$ \\
    Framework & PyTorch 2.11.0+cu126 \\
    Gaussian blur $\sigma$ & $\{1, 2, 4, 6, 8, 10, 14, 16, 20, 30, 40, 50, 60, 70\}$ \\
    Block size $B$ & $\{4, 8, 16, 38, 76, 152\}$ \\
    Hardware & $1\times$ NVIDIA GeForce RTX 4090 \\
    \bottomrule
  \end{tabular}
\end{table}
 
\begin{table}[tb]
  \centering
  \caption{Hyperparameters for Jigsaw Puzzle Solving.}
  \label{tab:hyperparams_exp3}
  
  \begin{tabular}{ll}
    \toprule
    \textbf{Hyperparameter} & \textbf{Value} \\
    \midrule
    Dataset & Animal Species Classification dataset \cite{Aashman2023_Animals} \\
    Puzzle sizes $k$ & $\{4, 6, 8\}$ \\
    Permutations & Sampled uniformly at random during training \\
    Data augmentation & None \\
    Model architecture & JPDVT \cite{Liu2024_JPDVT} \\
    Diffusion timesteps $T$ & $1000$ \\
    Model patch size & $16$ \\
    Initial weights & Random (trained from scratch) \\
    Input resolution & $3 \times 384 \times 384$ \\
    Loss function & Cosine similarity \\
    Optimizer & AdamW \cite{Loshchilov2019_AdamW} \\
    Learning rate & $10^{-4}$ \\
    Batch size & $64$ \\
    Max.\ epochs & $100$ \\
    Early stopping & patience 5 on validation loss \\
    Learning-rate schedule & Multistep at epochs $[20, 50, 75]$ with $\gamma = 0.1$ \\
    Checkpoint selection & Lowest validation loss \\
    Evaluation metric & Piece accuracy \\
    Random seed & $42$ \\
    Framework & PyTorch 2.11.0+cu126 \\
    Block size $B$ & Coupled to puzzle size $k$ \\
    Hardware & $2\times$ NVIDIA GeForce RTX 4090 \\
    \bottomrule
  \end{tabular}

\end{table}

\section*{Obfuscation-Utility Trade-off}

\begin{figure}[tb]
    \centering
    \includegraphics[width=\textwidth]{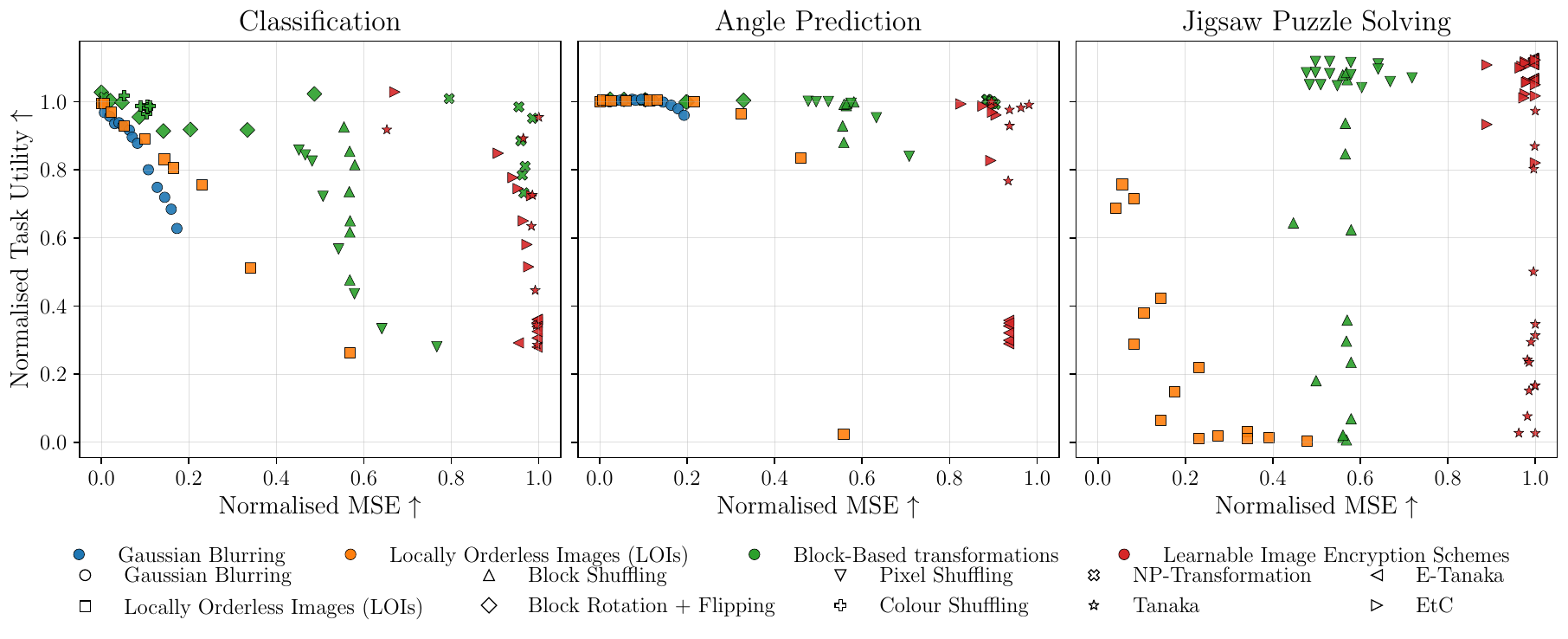}
    \caption{Obfuscation--utility trade-off: normalised task utility against normalised MSE.}
    \label{fig:UtilityVsSecurityMSE}
\end{figure}

\begin{figure}[tb]
    \centering
    \includegraphics[width=\textwidth]{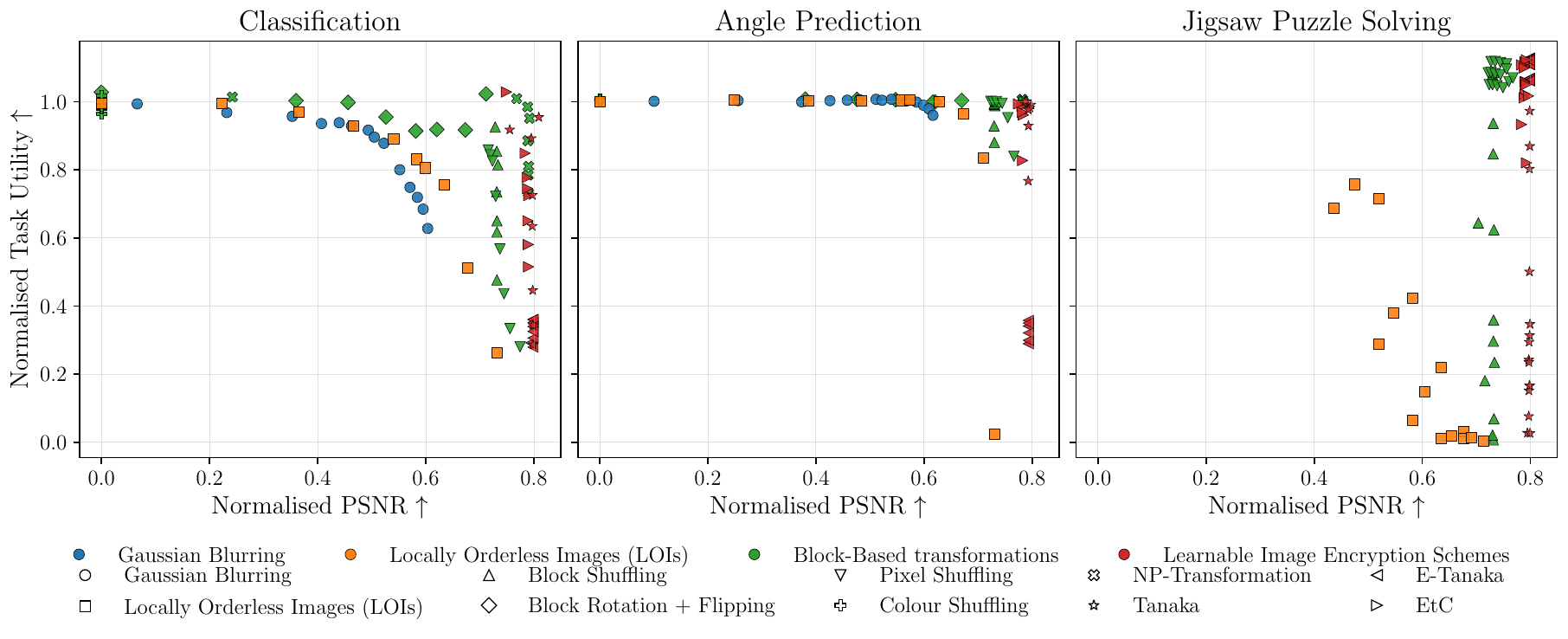}
    \caption{Obfuscation--utility trade-off: normalised task utility against normalised PSNR.}
    \label{fig:UtilityVsSecurityPSNR}
\end{figure}

\begin{figure}[tb]
    \centering
    \includegraphics[width=\textwidth]{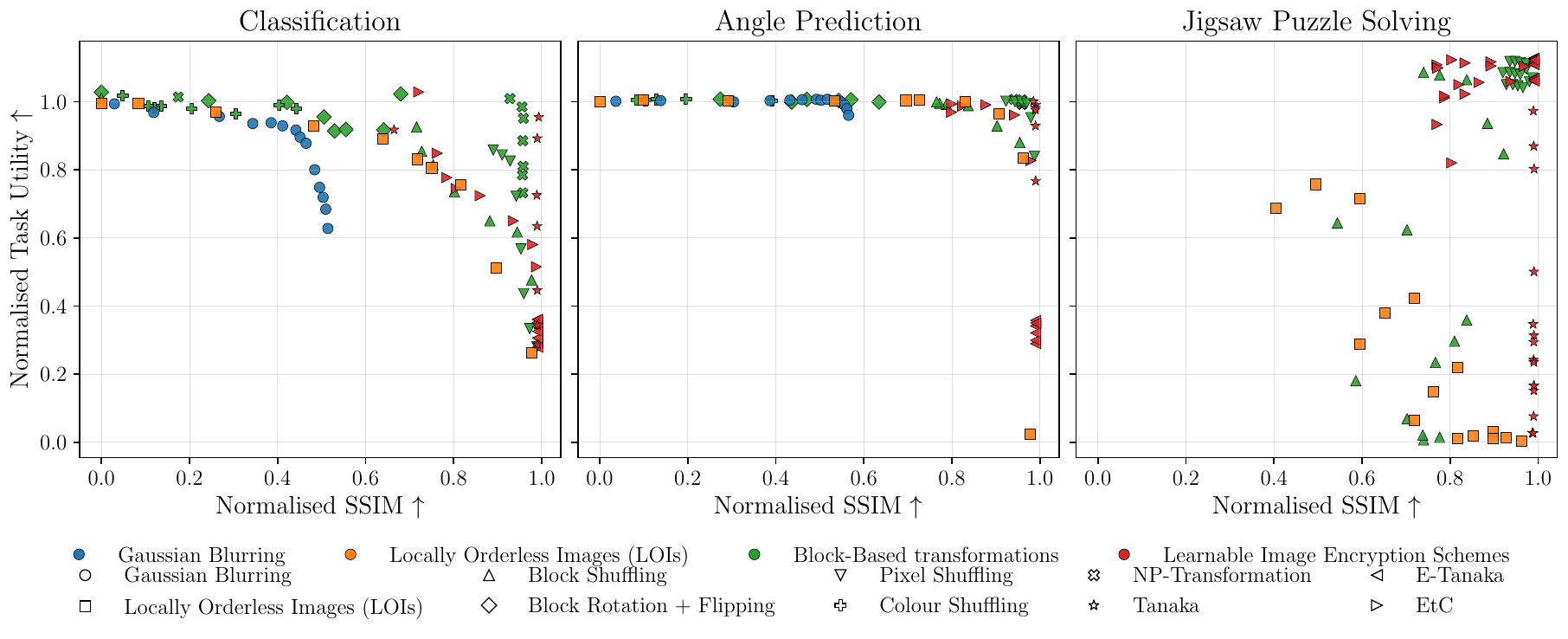}
    \caption{Obfuscation--utility trade-off: normalised task utility against normalised SSIM.}
    \label{fig:UtilityVsSecuritySSIM}
\end{figure}

\begin{figure}[tb]
    \centering
    \includegraphics[width=\textwidth]{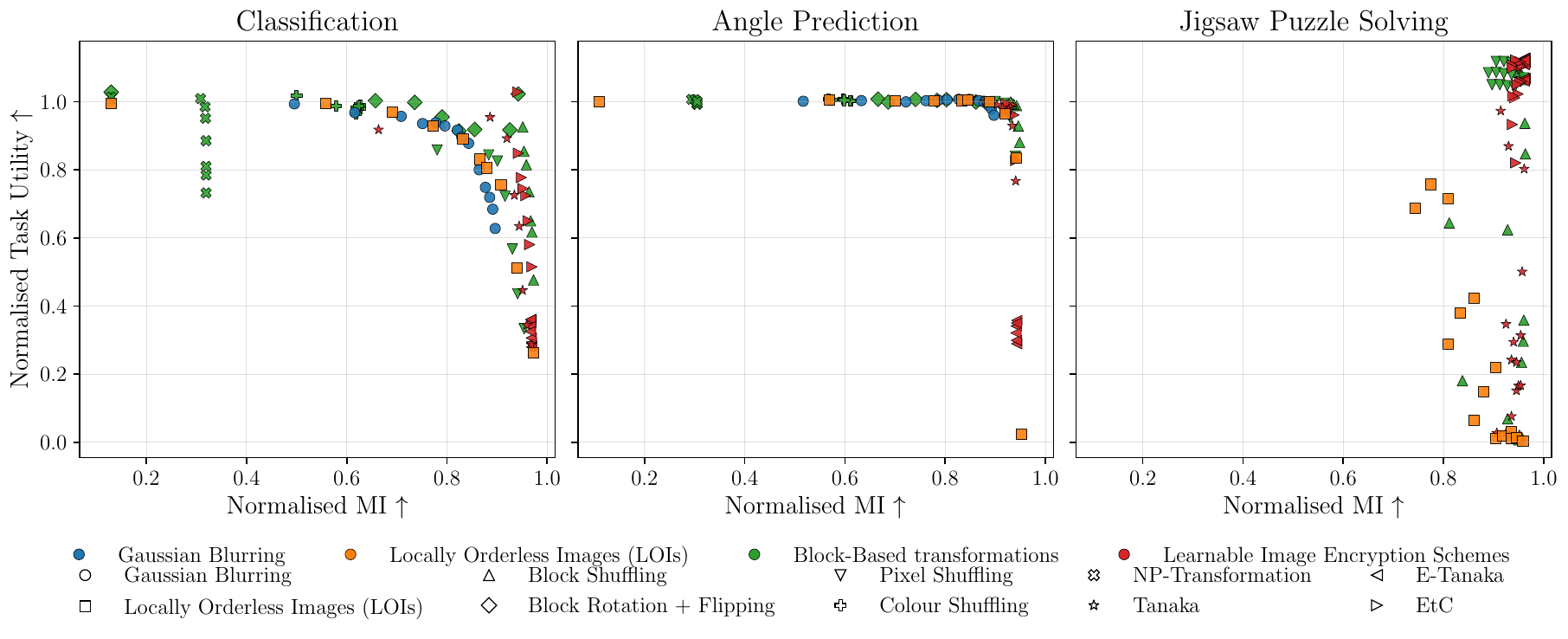}
    \caption{Obfuscation--utility trade-off: normalised task utility against normalised MI.}
    \label{fig:UtilityVsSecurityMI}
\end{figure}

\begin{figure}[tb]
    \centering
    \includegraphics[width=\textwidth]{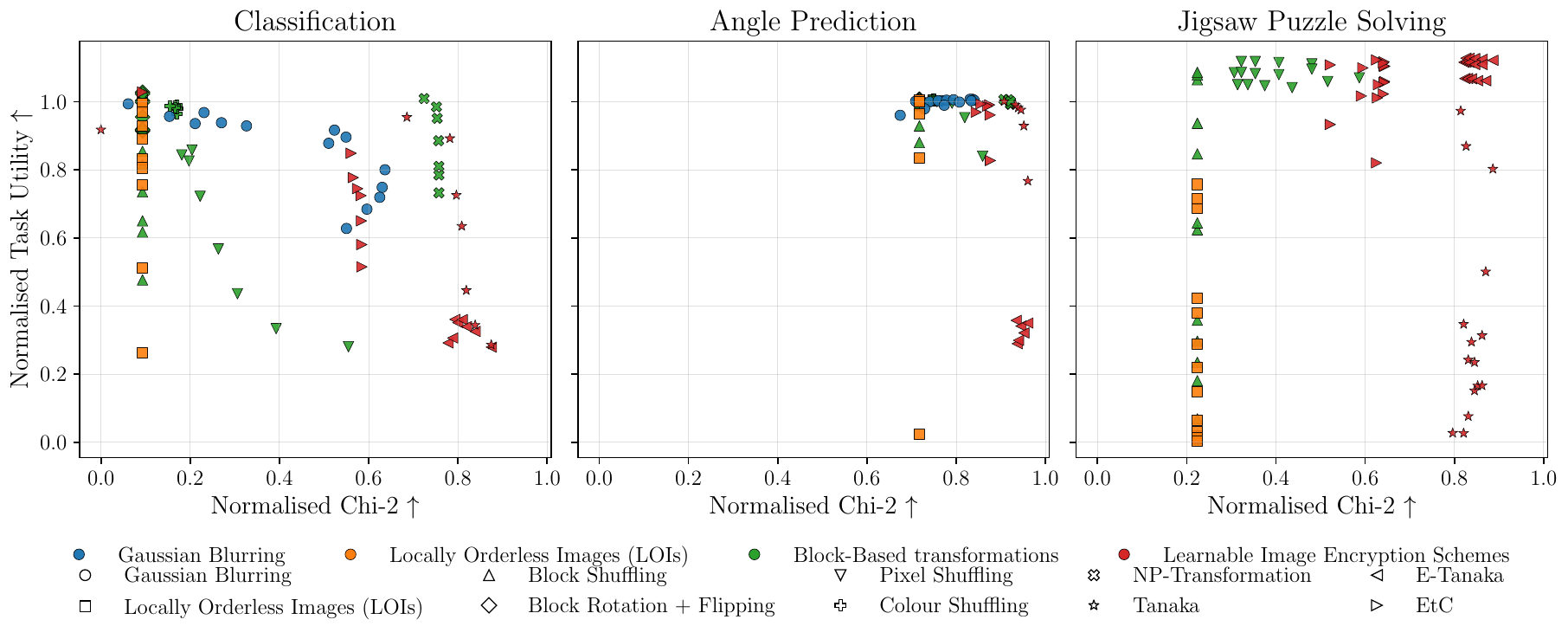}
    \caption{Obfuscation--utility trade-off: normalised task utility against normalised $\chi^2$.}
    \label{fig:UtilityVsSecurityChi2}
\end{figure}

\begin{figure}[tb]
    \centering
    \includegraphics[width=\textwidth]{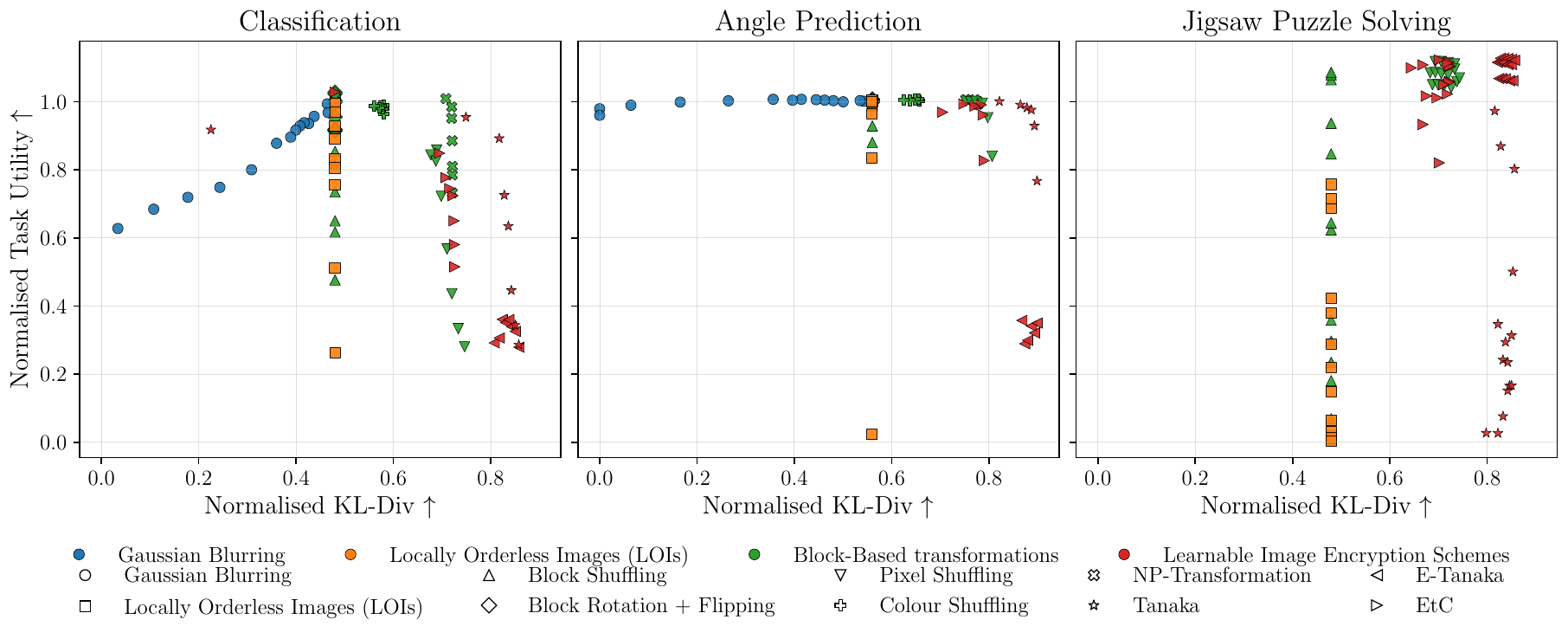}
    \caption{Obfuscation--utility trade-off: normalised task utility against normalised $D_{\mathrm{KL}}$.}
    \label{fig:UtilityVsSecurityKlDiv}
\end{figure}

\begin{figure}[tb]
    \centering
    \includegraphics[width=\textwidth]{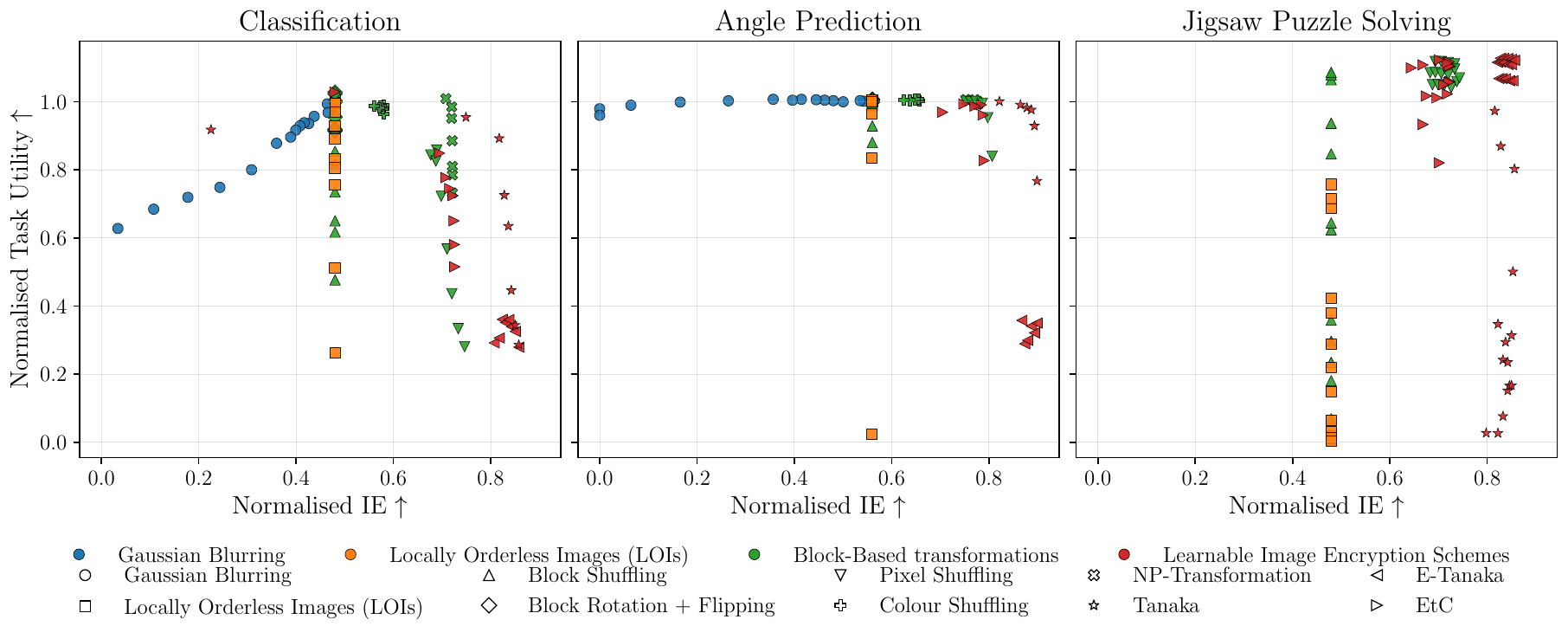}
    \caption{Obfuscation--utility trade-off: normalised task utility against normalised IE.}
    \label{fig:UtilityVsSecurityIE}
\end{figure}

\begin{figure}[tb]
    \centering
    \includegraphics[width=\textwidth]{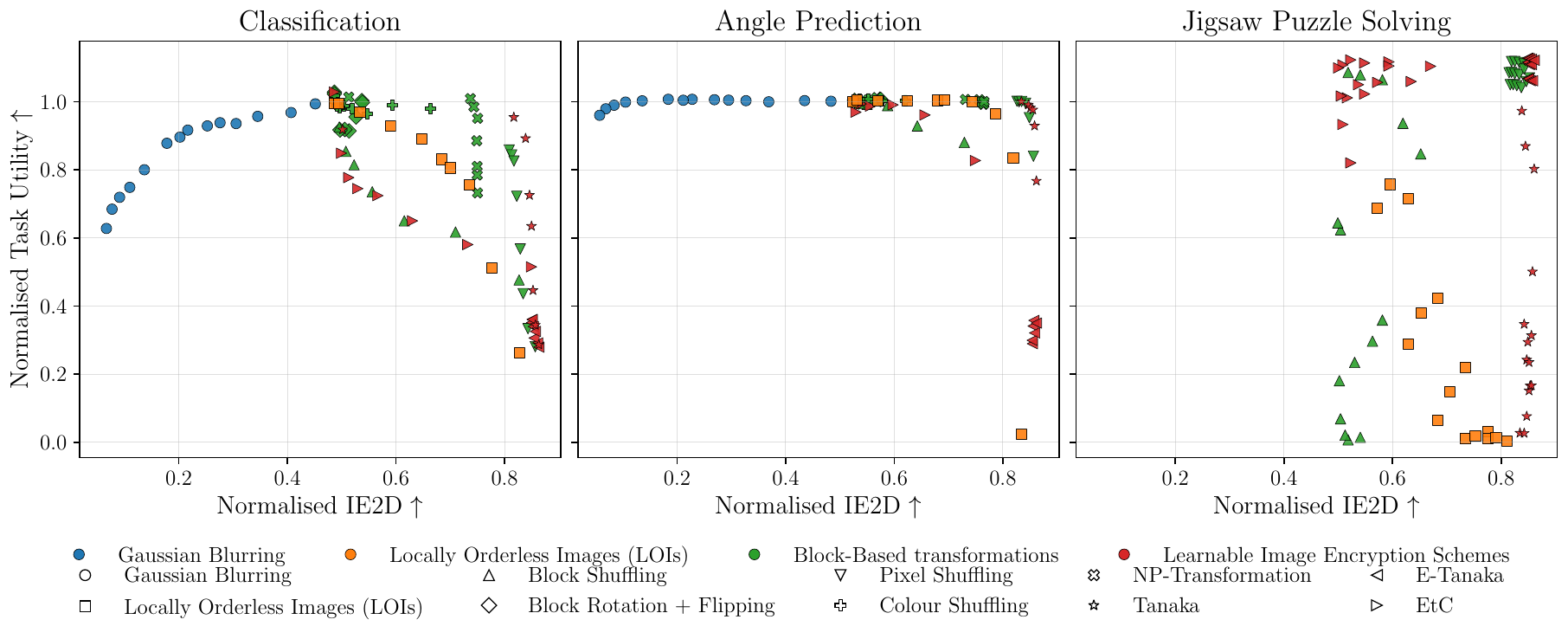}
    \caption{Obfuscation--utility trade-off: normalised task utility against normalised 2D-IE.}
    \label{fig:UtilityVsSecurityIE2D}
\end{figure}

\begin{figure}[tb]
    \centering
    \includegraphics[width=\textwidth]{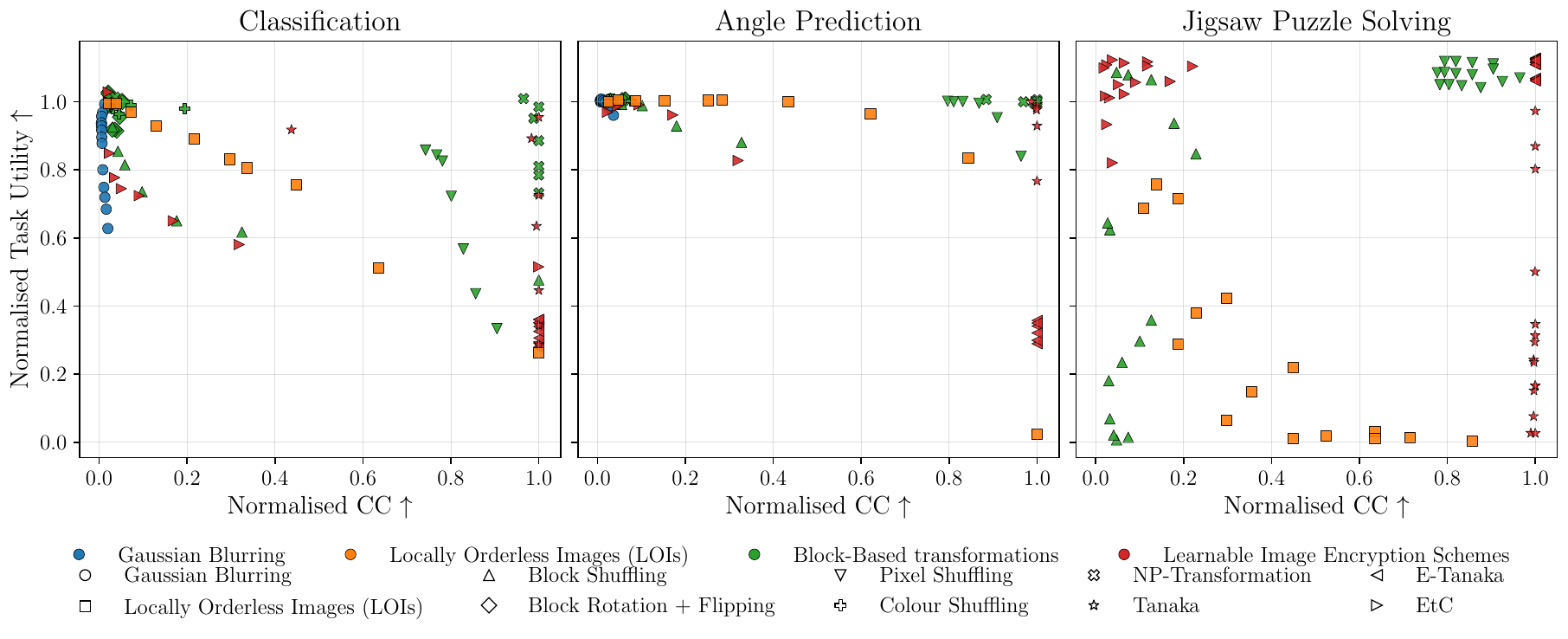}
    \caption{Obfuscation--utility trade-off: normalised task utility against normalised CC.}
    \label{fig:UtilityVsSecurityCC}
\end{figure}

\begin{figure}[tb]
    \centering
    \includegraphics[width=\textwidth]{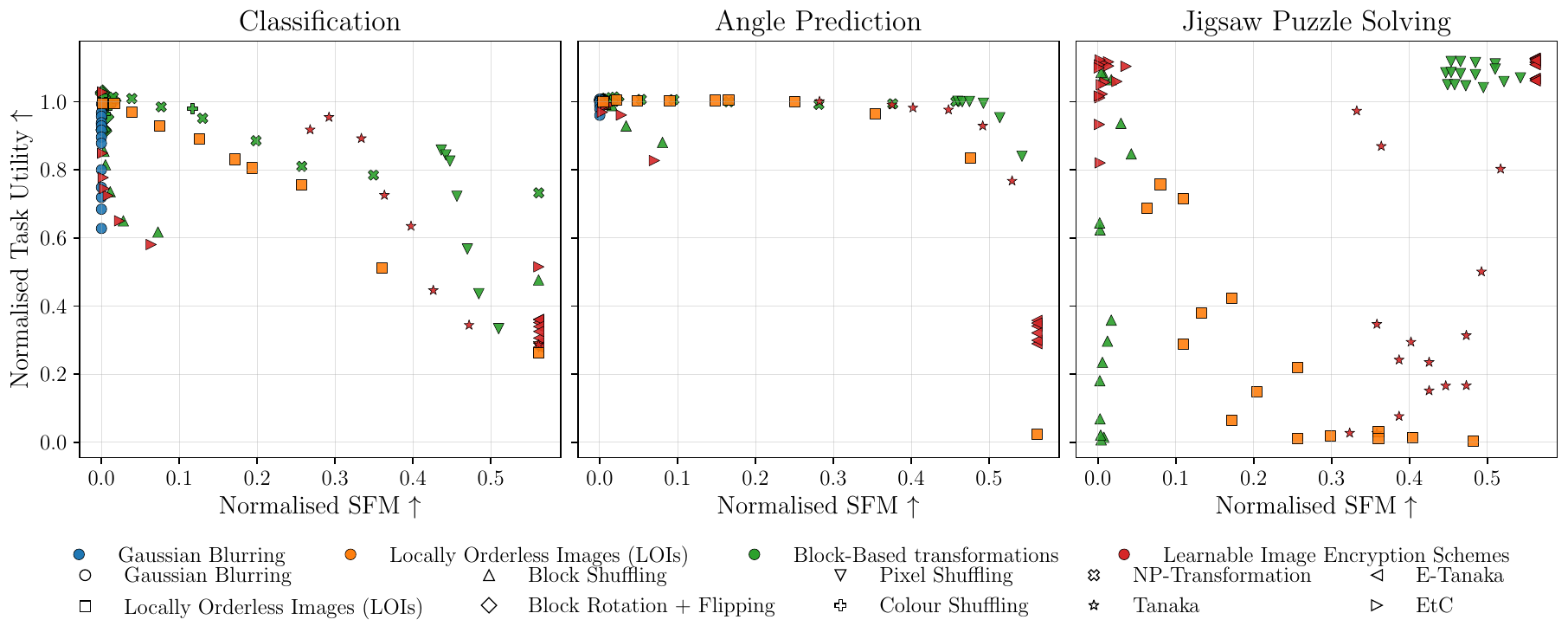}
    \caption{Obfuscation--utility trade-off: normalised task utility against normalised SFM.}
    \label{fig:UtilityVsSecuritySFM}
\end{figure}

\Cref{fig:UtilityVsSecurityMSE,fig:UtilityVsSecurityPSNR,fig:UtilityVsSecuritySSIM,fig:UtilityVsSecurityMI,fig:UtilityVsSecurityChi2,fig:UtilityVsSecurityKlDiv,fig:UtilityVsSecurityIE,fig:UtilityVsSecurityIE2D,fig:UtilityVsSecurityCC,fig:UtilityVsSecuritySFM} show the trade-off between task utility and the remaining image-domain obfuscation metrics complementing the LPIPS trade-off in the main paper. In each figure, colour indicates the transformation family and marker shape the individual transformation; each point corresponds to one transformation--scale configuration. Task utility is normalised per task as
\begin{equation*}
    u = \frac{\text{performance} - \text{chance}}{\text{baseline} - \text{chance}},
\end{equation*}
where \emph{baseline} denotes the plain-baseline performance and \emph{chance} the chance level ($1/N$ for classification, $90^\circ$ mean angular error for angle prediction, $1/k^2$ for jigsaw solving). For angle prediction, whose raw metric decreases with quality, the error is inverted before normalisation so that higher $u$ uniformly indicates higher utility. A value of $u = 1$ corresponds to plaintext-level performance and $u = 0$ to chance; values \emph{above} $1$ occur where transformed-domain performance exceeds the identity baseline. For keyed transformations, we attribute these values to deterministic key-induced shortcut cues. However, values above $1$ also occur for non-keyed transformations (notably angle prediction under mild Gaussian blur and small-block LOIs). In these cases the effect is explained by the transformation mildly aiding the specific task and model (\ie, low-pass filtering easing orientation estimation for a small CNN), rather than by shortcut memorisation.
 
The obfuscation metrics are min--max normalised per metric across all configurations, with direction-aware inversion so that higher values uniformly indicate stronger obfuscation. Differential and key-sensitivity metrics (NPCR, UACI, $|\mathcal{K}|$) are omitted from these plots, as they characterise a transformation's sensitivity rather than a per-configuration obfuscation level.

\end{document}

%% file: img/Examples.pgf
\begingroup%
\makeatletter%
\begin{pgfpicture}%
\pgfpathrectangle{\pgfpointorigin}{\pgfqpoint{4.998755in}{4.575693in}}%
\pgfusepath{use as bounding box, clip}%
\begin{pgfscope}%
\pgfsetbuttcap%
\pgfsetmiterjoin%
\definecolor{currentfill}{rgb}{1.000000,1.000000,1.000000}%
\pgfsetfillcolor{currentfill}%
\pgfsetlinewidth{0.000000pt}%
\definecolor{currentstroke}{rgb}{1.000000,1.000000,1.000000}%
\pgfsetstrokecolor{currentstroke}%
\pgfsetdash{}{0pt}%
\pgfpathmoveto{\pgfqpoint{0.000000in}{0.000000in}}%
\pgfpathlineto{\pgfqpoint{4.998755in}{0.000000in}}%
\pgfpathlineto{\pgfqpoint{4.998755in}{4.575693in}}%
\pgfpathlineto{\pgfqpoint{0.000000in}{4.575693in}}%
\pgfpathlineto{\pgfqpoint{0.000000in}{0.000000in}}%
\pgfpathclose%
\pgfusepath{fill}%
\end{pgfscope}%
\begin{pgfscope}%
\pgfpathrectangle{\pgfqpoint{0.100000in}{3.784674in}}{\pgfqpoint{0.548253in}{0.548253in}}%
\pgfusepath{clip}%
\pgfsys@transformcm{0.548253}{0.000000}{0.000000}{-0.548253}{0.100000in}{4.332928in}%
\pgftext[left,bottom]{\includegraphics[interpolate=false,width=1.000000in,height=1.000000in]{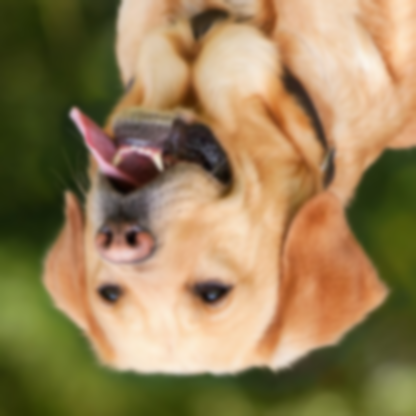}}%
\end{pgfscope}%
\begin{pgfscope}%
\definecolor{textcolor}{rgb}{0.000000,0.000000,0.000000}%
\pgfsetstrokecolor{textcolor}%
\pgfsetfillcolor{textcolor}%
\pgftext[x=0.374127in,y=3.751779in,,top]{\color{textcolor}{\rmfamily\fontsize{8.000000}{9.600000}\selectfont\catcode`\^=\active\def^{\ifmmode\sp\else\^{}\fi}\catcode`\%=\active\def
\end{pgfscope}%
\begin{pgfscope}%
\pgfpathrectangle{\pgfqpoint{0.692055in}{3.784674in}}{\pgfqpoint{0.548253in}{0.548253in}}%
\pgfusepath{clip}%
\pgfsys@transformcm{0.548253}{0.000000}{0.000000}{-0.548253}{0.692055in}{4.332928in}%
\pgftext[left,bottom]{\includegraphics[interpolate=false,width=1.000000in,height=1.000000in]{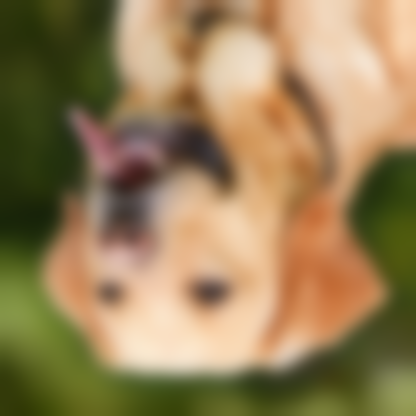}}%
\end{pgfscope}%
\begin{pgfscope}%
\definecolor{textcolor}{rgb}{0.000000,0.000000,0.000000}%
\pgfsetstrokecolor{textcolor}%
\pgfsetfillcolor{textcolor}%
\pgftext[x=0.966181in,y=3.751779in,,top]{\color{textcolor}{\rmfamily\fontsize{8.000000}{9.600000}\selectfont\catcode`\^=\active\def^{\ifmmode\sp\else\^{}\fi}\catcode`\%=\active\def
\end{pgfscope}%
\begin{pgfscope}%
\pgfpathrectangle{\pgfqpoint{1.284109in}{3.784674in}}{\pgfqpoint{0.548253in}{0.548253in}}%
\pgfusepath{clip}%
\pgfsys@transformcm{0.548253}{0.000000}{0.000000}{-0.548253}{1.284109in}{4.332928in}%
\pgftext[left,bottom]{\includegraphics[interpolate=false,width=1.000000in,height=1.000000in]{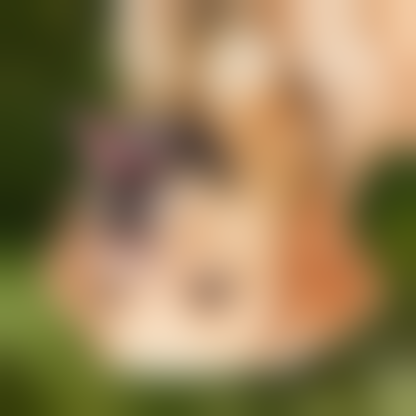}}%
\end{pgfscope}%
\begin{pgfscope}%
\definecolor{textcolor}{rgb}{0.000000,0.000000,0.000000}%
\pgfsetstrokecolor{textcolor}%
\pgfsetfillcolor{textcolor}%
\pgftext[x=1.558236in,y=3.751779in,,top]{\color{textcolor}{\rmfamily\fontsize{8.000000}{9.600000}\selectfont\catcode`\^=\active\def^{\ifmmode\sp\else\^{}\fi}\catcode`\%=\active\def
\end{pgfscope}%
\begin{pgfscope}%
\pgfpathrectangle{\pgfqpoint{1.876164in}{3.784674in}}{\pgfqpoint{0.548253in}{0.548253in}}%
\pgfusepath{clip}%
\pgfsys@transformcm{0.548253}{0.000000}{0.000000}{-0.548253}{1.876164in}{4.332928in}%
\pgftext[left,bottom]{\includegraphics[interpolate=false,width=1.000000in,height=1.000000in]{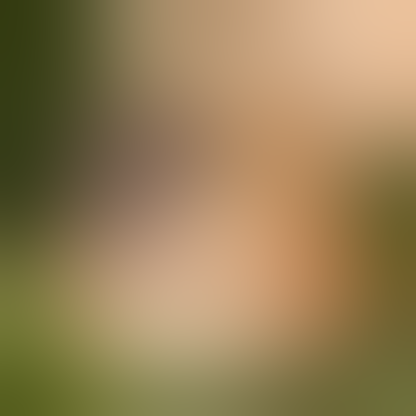}}%
\end{pgfscope}%
\begin{pgfscope}%
\definecolor{textcolor}{rgb}{0.000000,0.000000,0.000000}%
\pgfsetstrokecolor{textcolor}%
\pgfsetfillcolor{textcolor}%
\pgftext[x=2.150290in,y=3.751779in,,top]{\color{textcolor}{\rmfamily\fontsize{8.000000}{9.600000}\selectfont\catcode`\^=\active\def^{\ifmmode\sp\else\^{}\fi}\catcode`\%=\active\def
\end{pgfscope}%
\begin{pgfscope}%
\pgfpathrectangle{\pgfqpoint{2.574338in}{3.784674in}}{\pgfqpoint{0.548253in}{0.548253in}}%
\pgfusepath{clip}%
\pgfsys@transformcm{0.548253}{0.000000}{0.000000}{-0.548253}{2.574338in}{4.332928in}%
\pgftext[left,bottom]{\includegraphics[interpolate=false,width=1.000000in,height=1.000000in]{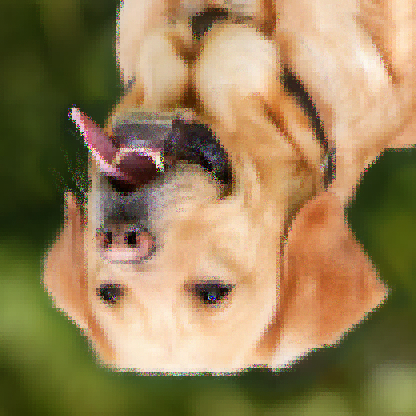}}%
\end{pgfscope}%
\begin{pgfscope}%
\definecolor{textcolor}{rgb}{0.000000,0.000000,0.000000}%
\pgfsetstrokecolor{textcolor}%
\pgfsetfillcolor{textcolor}%
\pgftext[x=2.848465in,y=3.751779in,,top]{\color{textcolor}{\rmfamily\fontsize{8.000000}{9.600000}\selectfont\catcode`\^=\active\def^{\ifmmode\sp\else\^{}\fi}\catcode`\%=\active\def
\end{pgfscope}%
\begin{pgfscope}%
\pgfpathrectangle{\pgfqpoint{3.166393in}{3.784674in}}{\pgfqpoint{0.548253in}{0.548253in}}%
\pgfusepath{clip}%
\pgfsys@transformcm{0.548253}{0.000000}{0.000000}{-0.548253}{3.166393in}{4.332928in}%
\pgftext[left,bottom]{\includegraphics[interpolate=false,width=1.000000in,height=1.000000in]{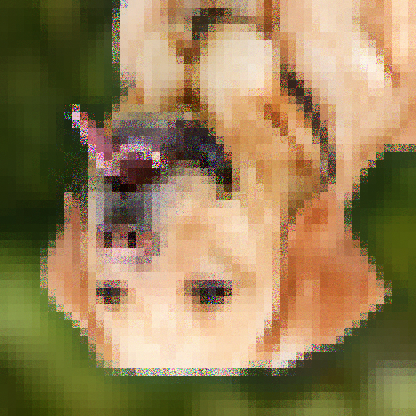}}%
\end{pgfscope}%
\begin{pgfscope}%
\definecolor{textcolor}{rgb}{0.000000,0.000000,0.000000}%
\pgfsetstrokecolor{textcolor}%
\pgfsetfillcolor{textcolor}%
\pgftext[x=3.440520in,y=3.751779in,,top]{\color{textcolor}{\rmfamily\fontsize{8.000000}{9.600000}\selectfont\catcode`\^=\active\def^{\ifmmode\sp\else\^{}\fi}\catcode`\%=\active\def
\end{pgfscope}%
\begin{pgfscope}%
\pgfpathrectangle{\pgfqpoint{3.758447in}{3.784674in}}{\pgfqpoint{0.548253in}{0.548253in}}%
\pgfusepath{clip}%
\pgfsys@transformcm{0.548253}{0.000000}{0.000000}{-0.548253}{3.758447in}{4.332928in}%
\pgftext[left,bottom]{\includegraphics[interpolate=false,width=1.000000in,height=1.000000in]{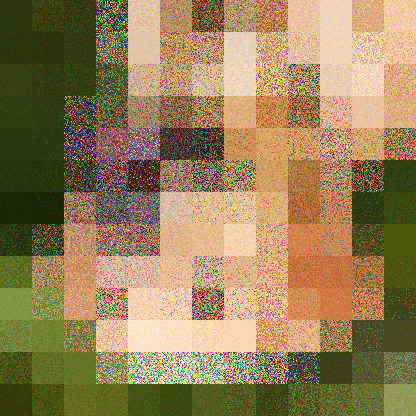}}%
\end{pgfscope}%
\begin{pgfscope}%
\definecolor{textcolor}{rgb}{0.000000,0.000000,0.000000}%
\pgfsetstrokecolor{textcolor}%
\pgfsetfillcolor{textcolor}%
\pgftext[x=4.032574in,y=3.751779in,,top]{\color{textcolor}{\rmfamily\fontsize{8.000000}{9.600000}\selectfont\catcode`\^=\active\def^{\ifmmode\sp\else\^{}\fi}\catcode`\%=\active\def
\end{pgfscope}%
\begin{pgfscope}%
\pgfpathrectangle{\pgfqpoint{4.350502in}{3.784674in}}{\pgfqpoint{0.548253in}{0.548253in}}%
\pgfusepath{clip}%
\pgfsys@transformcm{0.548253}{0.000000}{0.000000}{-0.548253}{4.350502in}{4.332928in}%
\pgftext[left,bottom]{\includegraphics[interpolate=false,width=1.000000in,height=1.000000in]{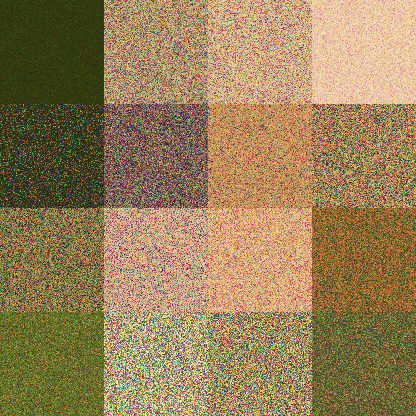}}%
\end{pgfscope}%
\begin{pgfscope}%
\definecolor{textcolor}{rgb}{0.000000,0.000000,0.000000}%
\pgfsetstrokecolor{textcolor}%
\pgfsetfillcolor{textcolor}%
\pgftext[x=4.624629in,y=3.751779in,,top]{\color{textcolor}{\rmfamily\fontsize{8.000000}{9.600000}\selectfont\catcode`\^=\active\def^{\ifmmode\sp\else\^{}\fi}\catcode`\%=\active\def
\end{pgfscope}%
\begin{pgfscope}%
\pgfpathrectangle{\pgfqpoint{0.100000in}{2.896421in}}{\pgfqpoint{0.548253in}{0.548253in}}%
\pgfusepath{clip}%
\pgfsys@transformcm{0.548253}{0.000000}{0.000000}{-0.548253}{0.100000in}{3.444674in}%
\pgftext[left,bottom]{\includegraphics[interpolate=false,width=1.000000in,height=1.000000in]{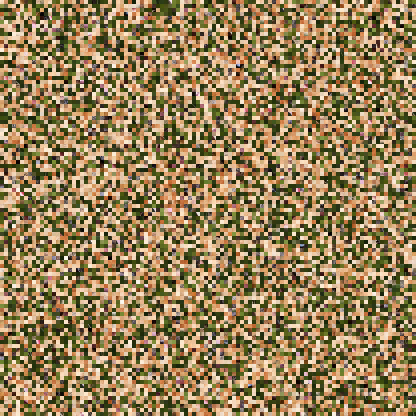}}%
\end{pgfscope}%
\begin{pgfscope}%
\definecolor{textcolor}{rgb}{0.000000,0.000000,0.000000}%
\pgfsetstrokecolor{textcolor}%
\pgfsetfillcolor{textcolor}%
\pgftext[x=0.374127in,y=2.863526in,,top]{\color{textcolor}{\rmfamily\fontsize{8.000000}{9.600000}\selectfont\catcode`\^=\active\def^{\ifmmode\sp\else\^{}\fi}\catcode`\%=\active\def
\end{pgfscope}%
\begin{pgfscope}%
\pgfpathrectangle{\pgfqpoint{0.692055in}{2.896421in}}{\pgfqpoint{0.548253in}{0.548253in}}%
\pgfusepath{clip}%
\pgfsys@transformcm{0.548253}{0.000000}{0.000000}{-0.548253}{0.692055in}{3.444674in}%
\pgftext[left,bottom]{\includegraphics[interpolate=false,width=1.000000in,height=1.000000in]{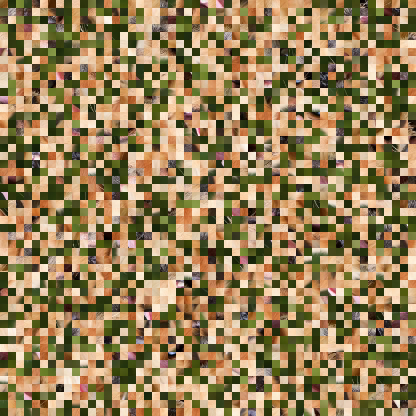}}%
\end{pgfscope}%
\begin{pgfscope}%
\definecolor{textcolor}{rgb}{0.000000,0.000000,0.000000}%
\pgfsetstrokecolor{textcolor}%
\pgfsetfillcolor{textcolor}%
\pgftext[x=0.966181in,y=2.863526in,,top]{\color{textcolor}{\rmfamily\fontsize{8.000000}{9.600000}\selectfont\catcode`\^=\active\def^{\ifmmode\sp\else\^{}\fi}\catcode`\%=\active\def
\end{pgfscope}%
\begin{pgfscope}%
\pgfpathrectangle{\pgfqpoint{1.284109in}{2.896421in}}{\pgfqpoint{0.548253in}{0.548253in}}%
\pgfusepath{clip}%
\pgfsys@transformcm{0.548253}{0.000000}{0.000000}{-0.548253}{1.284109in}{3.444674in}%
\pgftext[left,bottom]{\includegraphics[interpolate=false,width=1.000000in,height=1.000000in]{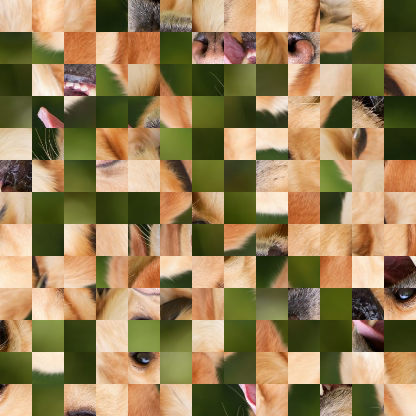}}%
\end{pgfscope}%
\begin{pgfscope}%
\definecolor{textcolor}{rgb}{0.000000,0.000000,0.000000}%
\pgfsetstrokecolor{textcolor}%
\pgfsetfillcolor{textcolor}%
\pgftext[x=1.558236in,y=2.863526in,,top]{\color{textcolor}{\rmfamily\fontsize{8.000000}{9.600000}\selectfont\catcode`\^=\active\def^{\ifmmode\sp\else\^{}\fi}\catcode`\%=\active\def
\end{pgfscope}%
\begin{pgfscope}%
\pgfpathrectangle{\pgfqpoint{1.876164in}{2.896421in}}{\pgfqpoint{0.548253in}{0.548253in}}%
\pgfusepath{clip}%
\pgfsys@transformcm{0.548253}{0.000000}{0.000000}{-0.548253}{1.876164in}{3.444674in}%
\pgftext[left,bottom]{\includegraphics[interpolate=false,width=1.000000in,height=1.000000in]{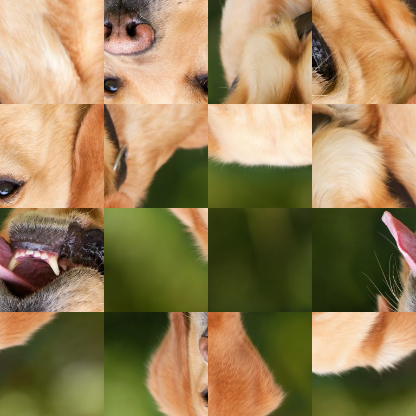}}%
\end{pgfscope}%
\begin{pgfscope}%
\definecolor{textcolor}{rgb}{0.000000,0.000000,0.000000}%
\pgfsetstrokecolor{textcolor}%
\pgfsetfillcolor{textcolor}%
\pgftext[x=2.150290in,y=2.863526in,,top]{\color{textcolor}{\rmfamily\fontsize{8.000000}{9.600000}\selectfont\catcode`\^=\active\def^{\ifmmode\sp\else\^{}\fi}\catcode`\%=\active\def
\end{pgfscope}%
\begin{pgfscope}%
\pgfpathrectangle{\pgfqpoint{2.574338in}{2.896421in}}{\pgfqpoint{0.548253in}{0.548253in}}%
\pgfusepath{clip}%
\pgfsys@transformcm{0.548253}{0.000000}{0.000000}{-0.548253}{2.574338in}{3.444674in}%
\pgftext[left,bottom]{\includegraphics[interpolate=false,width=1.000000in,height=1.000000in]{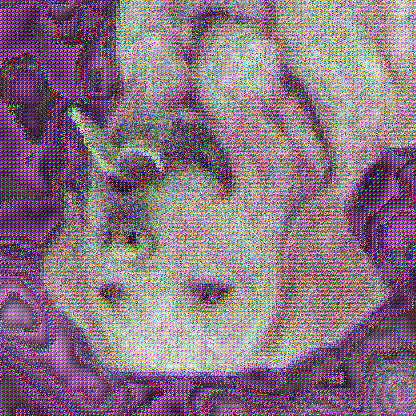}}%
\end{pgfscope}%
\begin{pgfscope}%
\definecolor{textcolor}{rgb}{0.000000,0.000000,0.000000}%
\pgfsetstrokecolor{textcolor}%
\pgfsetfillcolor{textcolor}%
\pgftext[x=2.848465in,y=2.863526in,,top]{\color{textcolor}{\rmfamily\fontsize{8.000000}{9.600000}\selectfont\catcode`\^=\active\def^{\ifmmode\sp\else\^{}\fi}\catcode`\%=\active\def
\end{pgfscope}%
\begin{pgfscope}%
\pgfpathrectangle{\pgfqpoint{3.166393in}{2.896421in}}{\pgfqpoint{0.548253in}{0.548253in}}%
\pgfusepath{clip}%
\pgfsys@transformcm{0.548253}{0.000000}{0.000000}{-0.548253}{3.166393in}{3.444674in}%
\pgftext[left,bottom]{\includegraphics[interpolate=false,width=1.000000in,height=1.000000in]{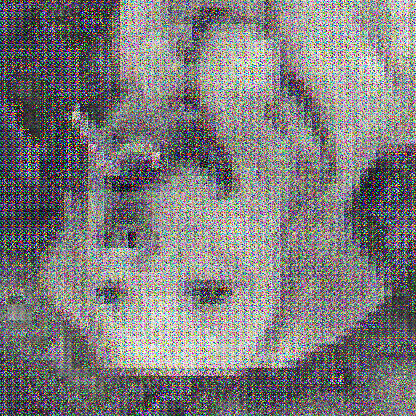}}%
\end{pgfscope}%
\begin{pgfscope}%
\definecolor{textcolor}{rgb}{0.000000,0.000000,0.000000}%
\pgfsetstrokecolor{textcolor}%
\pgfsetfillcolor{textcolor}%
\pgftext[x=3.440520in,y=2.863526in,,top]{\color{textcolor}{\rmfamily\fontsize{8.000000}{9.600000}\selectfont\catcode`\^=\active\def^{\ifmmode\sp\else\^{}\fi}\catcode`\%=\active\def
\end{pgfscope}%
\begin{pgfscope}%
\pgfpathrectangle{\pgfqpoint{3.758447in}{2.896421in}}{\pgfqpoint{0.548253in}{0.548253in}}%
\pgfusepath{clip}%
\pgfsys@transformcm{0.548253}{0.000000}{0.000000}{-0.548253}{3.758447in}{3.444674in}%
\pgftext[left,bottom]{\includegraphics[interpolate=false,width=1.000000in,height=1.000000in]{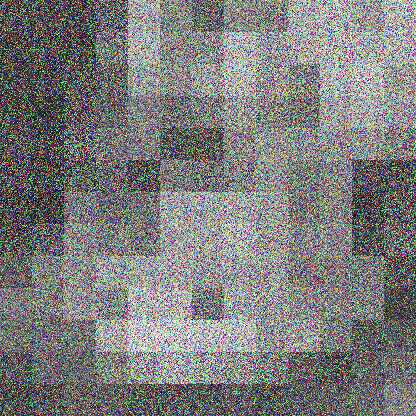}}%
\end{pgfscope}%
\begin{pgfscope}%
\definecolor{textcolor}{rgb}{0.000000,0.000000,0.000000}%
\pgfsetstrokecolor{textcolor}%
\pgfsetfillcolor{textcolor}%
\pgftext[x=4.032574in,y=2.863526in,,top]{\color{textcolor}{\rmfamily\fontsize{8.000000}{9.600000}\selectfont\catcode`\^=\active\def^{\ifmmode\sp\else\^{}\fi}\catcode`\%=\active\def
\end{pgfscope}%
\begin{pgfscope}%
\pgfpathrectangle{\pgfqpoint{4.350502in}{2.896421in}}{\pgfqpoint{0.548253in}{0.548253in}}%
\pgfusepath{clip}%
\pgfsys@transformcm{0.548253}{0.000000}{0.000000}{-0.548253}{4.350502in}{3.444674in}%
\pgftext[left,bottom]{\includegraphics[interpolate=false,width=1.000000in,height=1.000000in]{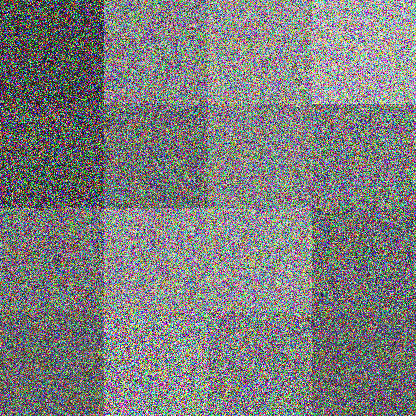}}%
\end{pgfscope}%
\begin{pgfscope}%
\definecolor{textcolor}{rgb}{0.000000,0.000000,0.000000}%
\pgfsetstrokecolor{textcolor}%
\pgfsetfillcolor{textcolor}%
\pgftext[x=4.624629in,y=2.863526in,,top]{\color{textcolor}{\rmfamily\fontsize{8.000000}{9.600000}\selectfont\catcode`\^=\active\def^{\ifmmode\sp\else\^{}\fi}\catcode`\%=\active\def
\end{pgfscope}%
\begin{pgfscope}%
\pgfpathrectangle{\pgfqpoint{0.100000in}{2.008168in}}{\pgfqpoint{0.548253in}{0.548253in}}%
\pgfusepath{clip}%
\pgfsys@transformcm{0.548253}{0.000000}{0.000000}{-0.548253}{0.100000in}{2.556421in}%
\pgftext[left,bottom]{\includegraphics[interpolate=false,width=1.000000in,height=1.000000in]{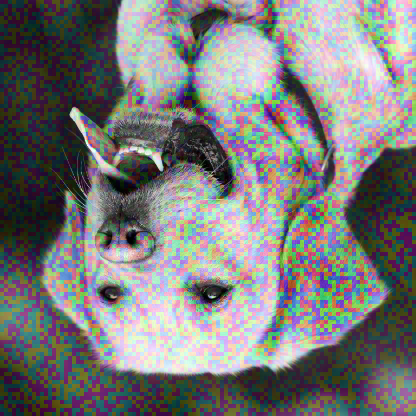}}%
\end{pgfscope}%
\begin{pgfscope}%
\definecolor{textcolor}{rgb}{0.000000,0.000000,0.000000}%
\pgfsetstrokecolor{textcolor}%
\pgfsetfillcolor{textcolor}%
\pgftext[x=0.374127in,y=1.975272in,,top]{\color{textcolor}{\rmfamily\fontsize{8.000000}{9.600000}\selectfont\catcode`\^=\active\def^{\ifmmode\sp\else\^{}\fi}\catcode`\%=\active\def
\end{pgfscope}%
\begin{pgfscope}%
\pgfpathrectangle{\pgfqpoint{0.692055in}{2.008168in}}{\pgfqpoint{0.548253in}{0.548253in}}%
\pgfusepath{clip}%
\pgfsys@transformcm{0.548253}{0.000000}{0.000000}{-0.548253}{0.692055in}{2.556421in}%
\pgftext[left,bottom]{\includegraphics[interpolate=false,width=1.000000in,height=1.000000in]{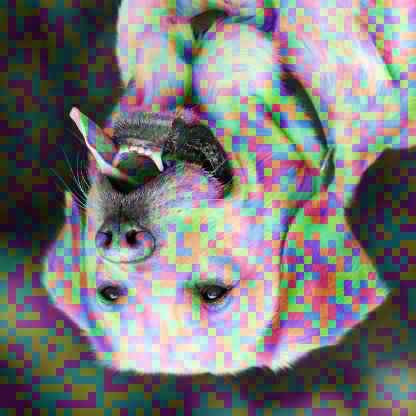}}%
\end{pgfscope}%
\begin{pgfscope}%
\definecolor{textcolor}{rgb}{0.000000,0.000000,0.000000}%
\pgfsetstrokecolor{textcolor}%
\pgfsetfillcolor{textcolor}%
\pgftext[x=0.966181in,y=1.975272in,,top]{\color{textcolor}{\rmfamily\fontsize{8.000000}{9.600000}\selectfont\catcode`\^=\active\def^{\ifmmode\sp\else\^{}\fi}\catcode`\%=\active\def
\end{pgfscope}%
\begin{pgfscope}%
\pgfpathrectangle{\pgfqpoint{1.284109in}{2.008168in}}{\pgfqpoint{0.548253in}{0.548253in}}%
\pgfusepath{clip}%
\pgfsys@transformcm{0.548253}{0.000000}{0.000000}{-0.548253}{1.284109in}{2.556421in}%
\pgftext[left,bottom]{\includegraphics[interpolate=false,width=1.000000in,height=1.000000in]{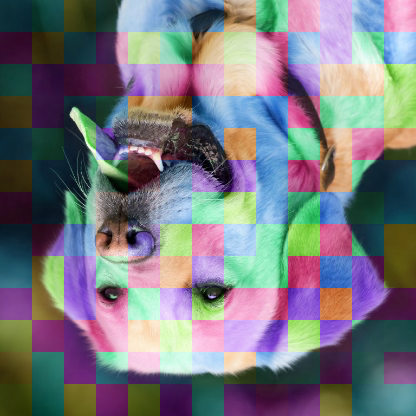}}%
\end{pgfscope}%
\begin{pgfscope}%
\definecolor{textcolor}{rgb}{0.000000,0.000000,0.000000}%
\pgfsetstrokecolor{textcolor}%
\pgfsetfillcolor{textcolor}%
\pgftext[x=1.558236in,y=1.975272in,,top]{\color{textcolor}{\rmfamily\fontsize{8.000000}{9.600000}\selectfont\catcode`\^=\active\def^{\ifmmode\sp\else\^{}\fi}\catcode`\%=\active\def
\end{pgfscope}%
\begin{pgfscope}%
\pgfpathrectangle{\pgfqpoint{1.876164in}{2.008168in}}{\pgfqpoint{0.548253in}{0.548253in}}%
\pgfusepath{clip}%
\pgfsys@transformcm{0.548253}{0.000000}{0.000000}{-0.548253}{1.876164in}{2.556421in}%
\pgftext[left,bottom]{\includegraphics[interpolate=false,width=1.000000in,height=1.000000in]{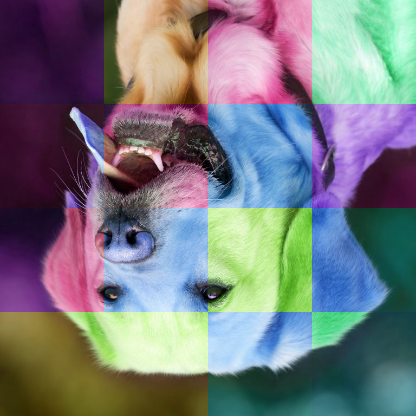}}%
\end{pgfscope}%
\begin{pgfscope}%
\definecolor{textcolor}{rgb}{0.000000,0.000000,0.000000}%
\pgfsetstrokecolor{textcolor}%
\pgfsetfillcolor{textcolor}%
\pgftext[x=2.150290in,y=1.975272in,,top]{\color{textcolor}{\rmfamily\fontsize{8.000000}{9.600000}\selectfont\catcode`\^=\active\def^{\ifmmode\sp\else\^{}\fi}\catcode`\%=\active\def
\end{pgfscope}%
\begin{pgfscope}%
\pgfpathrectangle{\pgfqpoint{2.574338in}{2.008168in}}{\pgfqpoint{0.548253in}{0.548253in}}%
\pgfusepath{clip}%
\pgfsys@transformcm{0.548253}{0.000000}{0.000000}{-0.548253}{2.574338in}{2.556421in}%
\pgftext[left,bottom]{\includegraphics[interpolate=false,width=1.000000in,height=1.000000in]{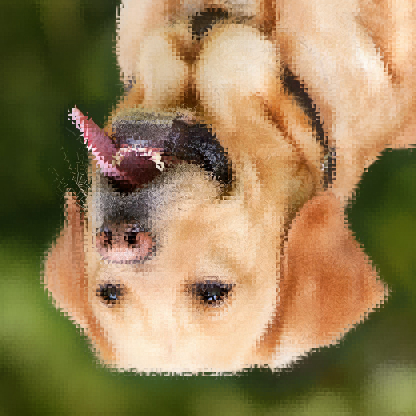}}%
\end{pgfscope}%
\begin{pgfscope}%
\definecolor{textcolor}{rgb}{0.000000,0.000000,0.000000}%
\pgfsetstrokecolor{textcolor}%
\pgfsetfillcolor{textcolor}%
\pgftext[x=2.848465in,y=1.975272in,,top]{\color{textcolor}{\rmfamily\fontsize{8.000000}{9.600000}\selectfont\catcode`\^=\active\def^{\ifmmode\sp\else\^{}\fi}\catcode`\%=\active\def
\end{pgfscope}%
\begin{pgfscope}%
\pgfpathrectangle{\pgfqpoint{3.166393in}{2.008168in}}{\pgfqpoint{0.548253in}{0.548253in}}%
\pgfusepath{clip}%
\pgfsys@transformcm{0.548253}{0.000000}{0.000000}{-0.548253}{3.166393in}{2.556421in}%
\pgftext[left,bottom]{\includegraphics[interpolate=false,width=1.000000in,height=1.000000in]{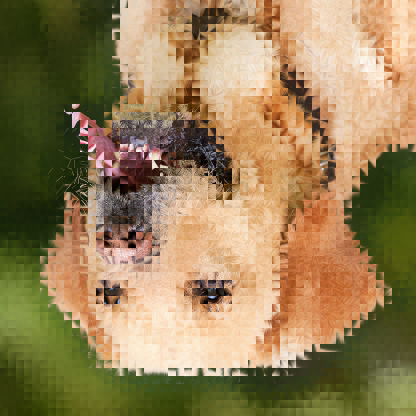}}%
\end{pgfscope}%
\begin{pgfscope}%
\definecolor{textcolor}{rgb}{0.000000,0.000000,0.000000}%
\pgfsetstrokecolor{textcolor}%
\pgfsetfillcolor{textcolor}%
\pgftext[x=3.440520in,y=1.975272in,,top]{\color{textcolor}{\rmfamily\fontsize{8.000000}{9.600000}\selectfont\catcode`\^=\active\def^{\ifmmode\sp\else\^{}\fi}\catcode`\%=\active\def
\end{pgfscope}%
\begin{pgfscope}%
\pgfpathrectangle{\pgfqpoint{3.758447in}{2.008168in}}{\pgfqpoint{0.548253in}{0.548253in}}%
\pgfusepath{clip}%
\pgfsys@transformcm{0.548253}{0.000000}{0.000000}{-0.548253}{3.758447in}{2.556421in}%
\pgftext[left,bottom]{\includegraphics[interpolate=false,width=1.000000in,height=1.000000in]{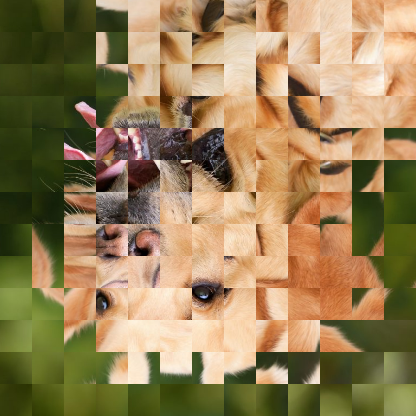}}%
\end{pgfscope}%
\begin{pgfscope}%
\definecolor{textcolor}{rgb}{0.000000,0.000000,0.000000}%
\pgfsetstrokecolor{textcolor}%
\pgfsetfillcolor{textcolor}%
\pgftext[x=4.032574in,y=1.975272in,,top]{\color{textcolor}{\rmfamily\fontsize{8.000000}{9.600000}\selectfont\catcode`\^=\active\def^{\ifmmode\sp\else\^{}\fi}\catcode`\%=\active\def
\end{pgfscope}%
\begin{pgfscope}%
\pgfpathrectangle{\pgfqpoint{4.350502in}{2.008168in}}{\pgfqpoint{0.548253in}{0.548253in}}%
\pgfusepath{clip}%
\pgfsys@transformcm{0.548253}{0.000000}{0.000000}{-0.548253}{4.350502in}{2.556421in}%
\pgftext[left,bottom]{\includegraphics[interpolate=false,width=1.000000in,height=1.000000in]{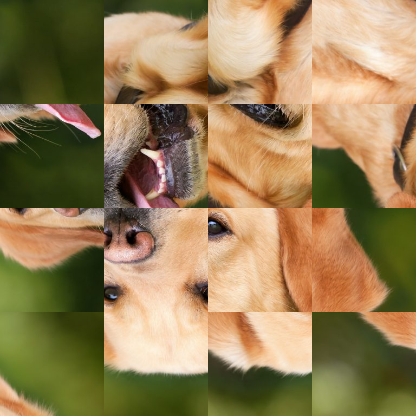}}%
\end{pgfscope}%
\begin{pgfscope}%
\definecolor{textcolor}{rgb}{0.000000,0.000000,0.000000}%
\pgfsetstrokecolor{textcolor}%
\pgfsetfillcolor{textcolor}%
\pgftext[x=4.624629in,y=1.975272in,,top]{\color{textcolor}{\rmfamily\fontsize{8.000000}{9.600000}\selectfont\catcode`\^=\active\def^{\ifmmode\sp\else\^{}\fi}\catcode`\%=\active\def
\end{pgfscope}%
\begin{pgfscope}%
\pgfpathrectangle{\pgfqpoint{0.100000in}{1.119914in}}{\pgfqpoint{0.548253in}{0.548253in}}%
\pgfusepath{clip}%
\pgfsys@transformcm{0.548253}{0.000000}{0.000000}{-0.548253}{0.100000in}{1.668168in}%
\pgftext[left,bottom]{\includegraphics[interpolate=false,width=1.000000in,height=1.000000in]{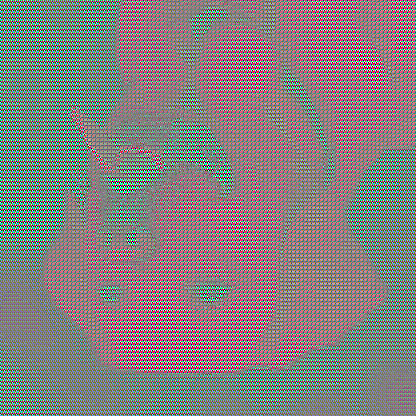}}%
\end{pgfscope}%
\begin{pgfscope}%
\definecolor{textcolor}{rgb}{0.000000,0.000000,0.000000}%
\pgfsetstrokecolor{textcolor}%
\pgfsetfillcolor{textcolor}%
\pgftext[x=0.374127in,y=1.087019in,,top]{\color{textcolor}{\rmfamily\fontsize{8.000000}{9.600000}\selectfont\catcode`\^=\active\def^{\ifmmode\sp\else\^{}\fi}\catcode`\%=\active\def
\end{pgfscope}%
\begin{pgfscope}%
\pgfpathrectangle{\pgfqpoint{0.692055in}{1.119914in}}{\pgfqpoint{0.548253in}{0.548253in}}%
\pgfusepath{clip}%
\pgfsys@transformcm{0.548253}{0.000000}{0.000000}{-0.548253}{0.692055in}{1.668168in}%
\pgftext[left,bottom]{\includegraphics[interpolate=false,width=1.000000in,height=1.000000in]{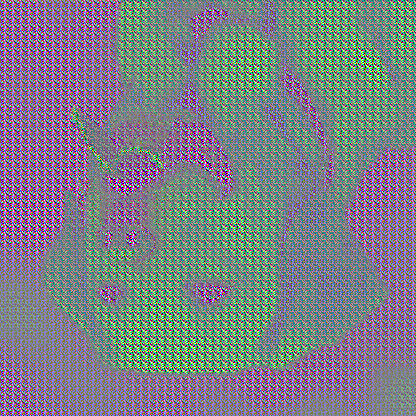}}%
\end{pgfscope}%
\begin{pgfscope}%
\definecolor{textcolor}{rgb}{0.000000,0.000000,0.000000}%
\pgfsetstrokecolor{textcolor}%
\pgfsetfillcolor{textcolor}%
\pgftext[x=0.966181in,y=1.087019in,,top]{\color{textcolor}{\rmfamily\fontsize{8.000000}{9.600000}\selectfont\catcode`\^=\active\def^{\ifmmode\sp\else\^{}\fi}\catcode`\%=\active\def
\end{pgfscope}%
\begin{pgfscope}%
\pgfpathrectangle{\pgfqpoint{1.284109in}{1.119914in}}{\pgfqpoint{0.548253in}{0.548253in}}%
\pgfusepath{clip}%
\pgfsys@transformcm{0.548253}{0.000000}{0.000000}{-0.548253}{1.284109in}{1.668168in}%
\pgftext[left,bottom]{\includegraphics[interpolate=false,width=1.000000in,height=1.000000in]{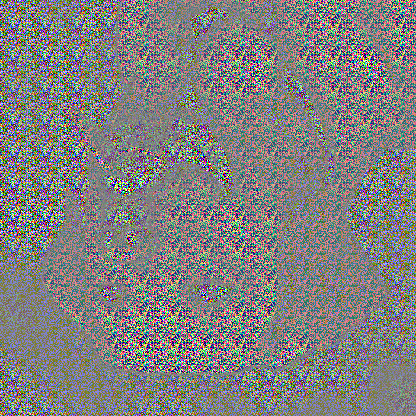}}%
\end{pgfscope}%
\begin{pgfscope}%
\definecolor{textcolor}{rgb}{0.000000,0.000000,0.000000}%
\pgfsetstrokecolor{textcolor}%
\pgfsetfillcolor{textcolor}%
\pgftext[x=1.558236in,y=1.087019in,,top]{\color{textcolor}{\rmfamily\fontsize{8.000000}{9.600000}\selectfont\catcode`\^=\active\def^{\ifmmode\sp\else\^{}\fi}\catcode`\%=\active\def
\end{pgfscope}%
\begin{pgfscope}%
\pgfpathrectangle{\pgfqpoint{1.876164in}{1.119914in}}{\pgfqpoint{0.548253in}{0.548253in}}%
\pgfusepath{clip}%
\pgfsys@transformcm{0.548253}{0.000000}{0.000000}{-0.548253}{1.876164in}{1.668168in}%
\pgftext[left,bottom]{\includegraphics[interpolate=false,width=1.000000in,height=1.000000in]{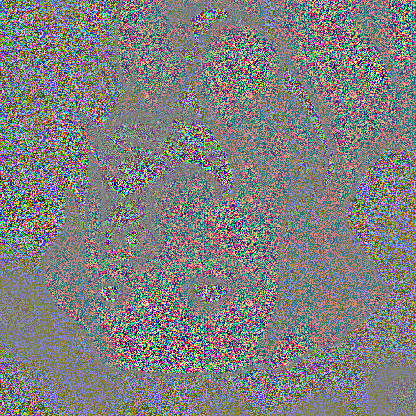}}%
\end{pgfscope}%
\begin{pgfscope}%
\definecolor{textcolor}{rgb}{0.000000,0.000000,0.000000}%
\pgfsetstrokecolor{textcolor}%
\pgfsetfillcolor{textcolor}%
\pgftext[x=2.150290in,y=1.087019in,,top]{\color{textcolor}{\rmfamily\fontsize{8.000000}{9.600000}\selectfont\catcode`\^=\active\def^{\ifmmode\sp\else\^{}\fi}\catcode`\%=\active\def
\end{pgfscope}%
\begin{pgfscope}%
\pgfpathrectangle{\pgfqpoint{2.574338in}{1.119914in}}{\pgfqpoint{0.548253in}{0.548253in}}%
\pgfusepath{clip}%
\pgfsys@transformcm{0.548253}{0.000000}{0.000000}{-0.548253}{2.574338in}{1.668168in}%
\pgftext[left,bottom]{\includegraphics[interpolate=false,width=1.000000in,height=1.000000in]{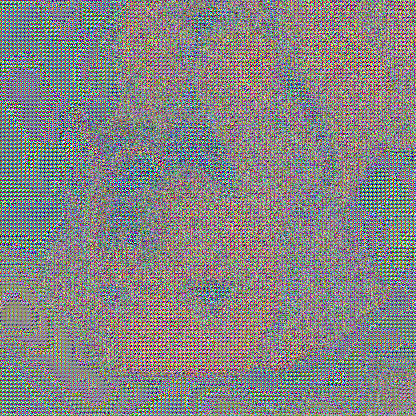}}%
\end{pgfscope}%
\begin{pgfscope}%
\definecolor{textcolor}{rgb}{0.000000,0.000000,0.000000}%
\pgfsetstrokecolor{textcolor}%
\pgfsetfillcolor{textcolor}%
\pgftext[x=2.848465in,y=1.087019in,,top]{\color{textcolor}{\rmfamily\fontsize{8.000000}{9.600000}\selectfont\catcode`\^=\active\def^{\ifmmode\sp\else\^{}\fi}\catcode`\%=\active\def
\end{pgfscope}%
\begin{pgfscope}%
\pgfpathrectangle{\pgfqpoint{3.166393in}{1.119914in}}{\pgfqpoint{0.548253in}{0.548253in}}%
\pgfusepath{clip}%
\pgfsys@transformcm{0.548253}{0.000000}{0.000000}{-0.548253}{3.166393in}{1.668168in}%
\pgftext[left,bottom]{\includegraphics[interpolate=false,width=1.000000in,height=1.000000in]{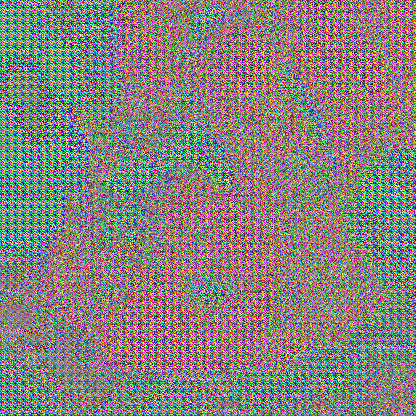}}%
\end{pgfscope}%
\begin{pgfscope}%
\definecolor{textcolor}{rgb}{0.000000,0.000000,0.000000}%
\pgfsetstrokecolor{textcolor}%
\pgfsetfillcolor{textcolor}%
\pgftext[x=3.440520in,y=1.087019in,,top]{\color{textcolor}{\rmfamily\fontsize{8.000000}{9.600000}\selectfont\catcode`\^=\active\def^{\ifmmode\sp\else\^{}\fi}\catcode`\%=\active\def
\end{pgfscope}%
\begin{pgfscope}%
\pgfpathrectangle{\pgfqpoint{3.758447in}{1.119914in}}{\pgfqpoint{0.548253in}{0.548253in}}%
\pgfusepath{clip}%
\pgfsys@transformcm{0.548253}{0.000000}{0.000000}{-0.548253}{3.758447in}{1.668168in}%
\pgftext[left,bottom]{\includegraphics[interpolate=false,width=1.000000in,height=1.000000in]{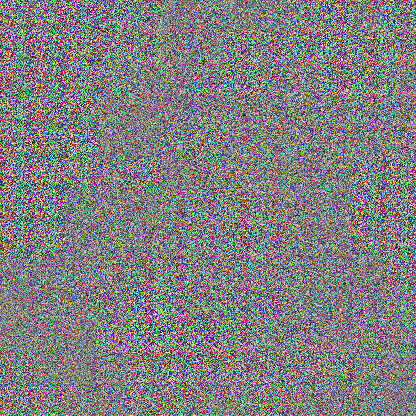}}%
\end{pgfscope}%
\begin{pgfscope}%
\definecolor{textcolor}{rgb}{0.000000,0.000000,0.000000}%
\pgfsetstrokecolor{textcolor}%
\pgfsetfillcolor{textcolor}%
\pgftext[x=4.032574in,y=1.087019in,,top]{\color{textcolor}{\rmfamily\fontsize{8.000000}{9.600000}\selectfont\catcode`\^=\active\def^{\ifmmode\sp\else\^{}\fi}\catcode`\%=\active\def
\end{pgfscope}%
\begin{pgfscope}%
\pgfpathrectangle{\pgfqpoint{4.350502in}{1.119914in}}{\pgfqpoint{0.548253in}{0.548253in}}%
\pgfusepath{clip}%
\pgfsys@transformcm{0.548253}{0.000000}{0.000000}{-0.548253}{4.350502in}{1.668168in}%
\pgftext[left,bottom]{\includegraphics[interpolate=false,width=1.000000in,height=1.000000in]{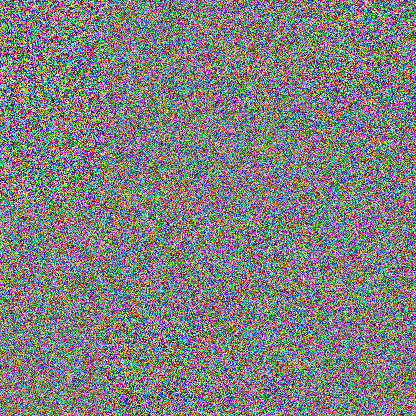}}%
\end{pgfscope}%
\begin{pgfscope}%
\definecolor{textcolor}{rgb}{0.000000,0.000000,0.000000}%
\pgfsetstrokecolor{textcolor}%
\pgfsetfillcolor{textcolor}%
\pgftext[x=4.624629in,y=1.087019in,,top]{\color{textcolor}{\rmfamily\fontsize{8.000000}{9.600000}\selectfont\catcode`\^=\active\def^{\ifmmode\sp\else\^{}\fi}\catcode`\%=\active\def
\end{pgfscope}%
\begin{pgfscope}%
\pgfpathrectangle{\pgfqpoint{0.100000in}{0.231661in}}{\pgfqpoint{0.548253in}{0.548253in}}%
\pgfusepath{clip}%
\pgfsys@transformcm{0.548253}{0.000000}{0.000000}{-0.548253}{0.100000in}{0.779914in}%
\pgftext[left,bottom]{\includegraphics[interpolate=false,width=1.000000in,height=1.000000in]{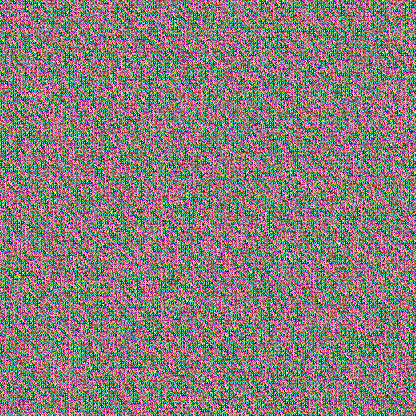}}%
\end{pgfscope}%
\begin{pgfscope}%
\definecolor{textcolor}{rgb}{0.000000,0.000000,0.000000}%
\pgfsetstrokecolor{textcolor}%
\pgfsetfillcolor{textcolor}%
\pgftext[x=0.374127in,y=0.198766in,,top]{\color{textcolor}{\rmfamily\fontsize{8.000000}{9.600000}\selectfont\catcode`\^=\active\def^{\ifmmode\sp\else\^{}\fi}\catcode`\%=\active\def
\end{pgfscope}%
\begin{pgfscope}%
\pgfpathrectangle{\pgfqpoint{0.692055in}{0.231661in}}{\pgfqpoint{0.548253in}{0.548253in}}%
\pgfusepath{clip}%
\pgfsys@transformcm{0.548253}{0.000000}{0.000000}{-0.548253}{0.692055in}{0.779914in}%
\pgftext[left,bottom]{\includegraphics[interpolate=false,width=1.000000in,height=1.000000in]{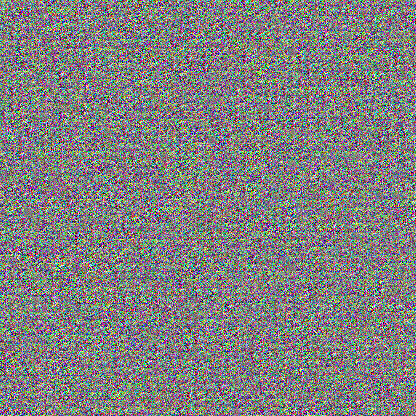}}%
\end{pgfscope}%
\begin{pgfscope}%
\definecolor{textcolor}{rgb}{0.000000,0.000000,0.000000}%
\pgfsetstrokecolor{textcolor}%
\pgfsetfillcolor{textcolor}%
\pgftext[x=0.966181in,y=0.198766in,,top]{\color{textcolor}{\rmfamily\fontsize{8.000000}{9.600000}\selectfont\catcode`\^=\active\def^{\ifmmode\sp\else\^{}\fi}\catcode`\%=\active\def
\end{pgfscope}%
\begin{pgfscope}%
\pgfpathrectangle{\pgfqpoint{1.284109in}{0.231661in}}{\pgfqpoint{0.548253in}{0.548253in}}%
\pgfusepath{clip}%
\pgfsys@transformcm{0.548253}{0.000000}{0.000000}{-0.548253}{1.284109in}{0.779914in}%
\pgftext[left,bottom]{\includegraphics[interpolate=false,width=1.000000in,height=1.000000in]{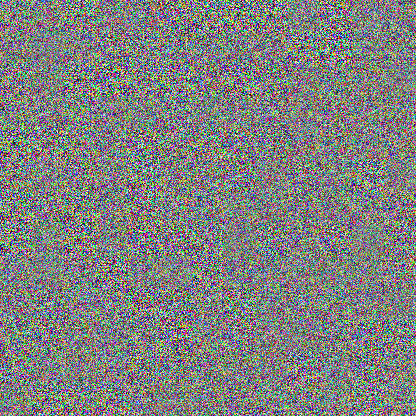}}%
\end{pgfscope}%
\begin{pgfscope}%
\definecolor{textcolor}{rgb}{0.000000,0.000000,0.000000}%
\pgfsetstrokecolor{textcolor}%
\pgfsetfillcolor{textcolor}%
\pgftext[x=1.558236in,y=0.198766in,,top]{\color{textcolor}{\rmfamily\fontsize{8.000000}{9.600000}\selectfont\catcode`\^=\active\def^{\ifmmode\sp\else\^{}\fi}\catcode`\%=\active\def
\end{pgfscope}%
\begin{pgfscope}%
\pgfpathrectangle{\pgfqpoint{1.876164in}{0.231661in}}{\pgfqpoint{0.548253in}{0.548253in}}%
\pgfusepath{clip}%
\pgfsys@transformcm{0.548253}{0.000000}{0.000000}{-0.548253}{1.876164in}{0.779914in}%
\pgftext[left,bottom]{\includegraphics[interpolate=false,width=1.000000in,height=1.000000in]{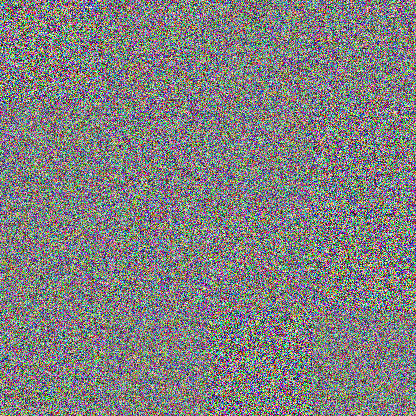}}%
\end{pgfscope}%
\begin{pgfscope}%
\definecolor{textcolor}{rgb}{0.000000,0.000000,0.000000}%
\pgfsetstrokecolor{textcolor}%
\pgfsetfillcolor{textcolor}%
\pgftext[x=2.150290in,y=0.198766in,,top]{\color{textcolor}{\rmfamily\fontsize{8.000000}{9.600000}\selectfont\catcode`\^=\active\def^{\ifmmode\sp\else\^{}\fi}\catcode`\%=\active\def
\end{pgfscope}%
\begin{pgfscope}%
\pgfpathrectangle{\pgfqpoint{2.574338in}{0.231661in}}{\pgfqpoint{0.548253in}{0.548253in}}%
\pgfusepath{clip}%
\pgfsys@transformcm{0.548253}{0.000000}{0.000000}{-0.548253}{2.574338in}{0.779914in}%
\pgftext[left,bottom]{\includegraphics[interpolate=false,width=1.000000in,height=1.000000in]{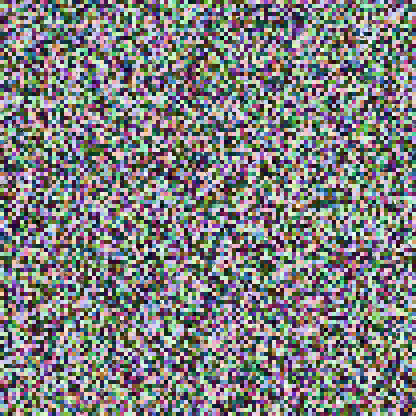}}%
\end{pgfscope}%
\begin{pgfscope}%
\definecolor{textcolor}{rgb}{0.000000,0.000000,0.000000}%
\pgfsetstrokecolor{textcolor}%
\pgfsetfillcolor{textcolor}%
\pgftext[x=2.848465in,y=0.198766in,,top]{\color{textcolor}{\rmfamily\fontsize{8.000000}{9.600000}\selectfont\catcode`\^=\active\def^{\ifmmode\sp\else\^{}\fi}\catcode`\%=\active\def
\end{pgfscope}%
\begin{pgfscope}%
\pgfpathrectangle{\pgfqpoint{3.166393in}{0.231661in}}{\pgfqpoint{0.548253in}{0.548253in}}%
\pgfusepath{clip}%
\pgfsys@transformcm{0.548253}{0.000000}{0.000000}{-0.548253}{3.166393in}{0.779914in}%
\pgftext[left,bottom]{\includegraphics[interpolate=false,width=1.000000in,height=1.000000in]{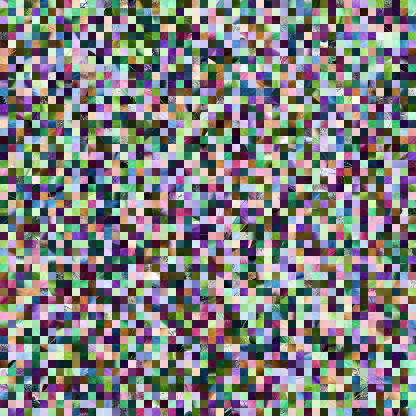}}%
\end{pgfscope}%
\begin{pgfscope}%
\definecolor{textcolor}{rgb}{0.000000,0.000000,0.000000}%
\pgfsetstrokecolor{textcolor}%
\pgfsetfillcolor{textcolor}%
\pgftext[x=3.440520in,y=0.198766in,,top]{\color{textcolor}{\rmfamily\fontsize{8.000000}{9.600000}\selectfont\catcode`\^=\active\def^{\ifmmode\sp\else\^{}\fi}\catcode`\%=\active\def
\end{pgfscope}%
\begin{pgfscope}%
\pgfpathrectangle{\pgfqpoint{3.758447in}{0.231661in}}{\pgfqpoint{0.548253in}{0.548253in}}%
\pgfusepath{clip}%
\pgfsys@transformcm{0.548253}{0.000000}{0.000000}{-0.548253}{3.758447in}{0.779914in}%
\pgftext[left,bottom]{\includegraphics[interpolate=false,width=1.000000in,height=1.000000in]{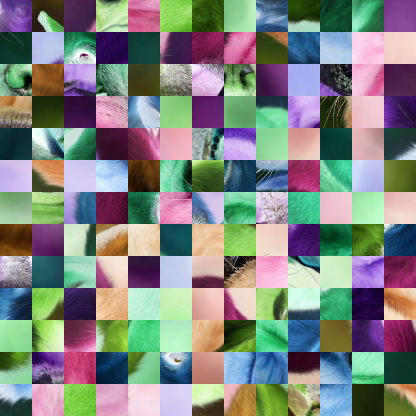}}%
\end{pgfscope}%
\begin{pgfscope}%
\definecolor{textcolor}{rgb}{0.000000,0.000000,0.000000}%
\pgfsetstrokecolor{textcolor}%
\pgfsetfillcolor{textcolor}%
\pgftext[x=4.032574in,y=0.198766in,,top]{\color{textcolor}{\rmfamily\fontsize{8.000000}{9.600000}\selectfont\catcode`\^=\active\def^{\ifmmode\sp\else\^{}\fi}\catcode`\%=\active\def
\end{pgfscope}%
\begin{pgfscope}%
\pgfpathrectangle{\pgfqpoint{4.350502in}{0.231661in}}{\pgfqpoint{0.548253in}{0.548253in}}%
\pgfusepath{clip}%
\pgfsys@transformcm{0.548253}{0.000000}{0.000000}{-0.548253}{4.350502in}{0.779914in}%
\pgftext[left,bottom]{\includegraphics[interpolate=false,width=1.000000in,height=1.000000in]{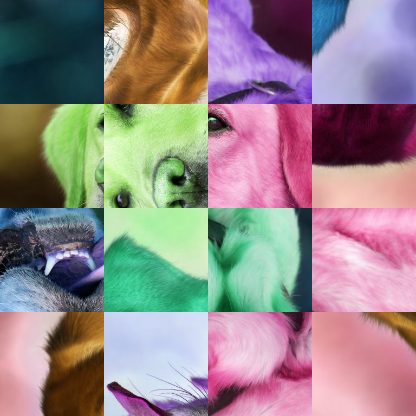}}%
\end{pgfscope}%
\begin{pgfscope}%
\definecolor{textcolor}{rgb}{0.000000,0.000000,0.000000}%
\pgfsetstrokecolor{textcolor}%
\pgfsetfillcolor{textcolor}%
\pgftext[x=4.624629in,y=0.198766in,,top]{\color{textcolor}{\rmfamily\fontsize{8.000000}{9.600000}\selectfont\catcode`\^=\active\def^{\ifmmode\sp\else\^{}\fi}\catcode`\%=\active\def
\end{pgfscope}%
\begin{pgfscope}%
\definecolor{textcolor}{rgb}{0.000000,0.000000,0.000000}%
\pgfsetstrokecolor{textcolor}%
\pgfsetfillcolor{textcolor}%
\pgftext[x=1.262209in,y=4.350693in,,bottom]{\color{textcolor}{\rmfamily\fontsize{9.000000}{10.800000}\selectfont\catcode`\^=\active\def^{\ifmmode\sp\else\^{}\fi}\catcode`\%=\active\def
\end{pgfscope}%
\begin{pgfscope}%
\definecolor{textcolor}{rgb}{0.000000,0.000000,0.000000}%
\pgfsetstrokecolor{textcolor}%
\pgfsetfillcolor{textcolor}%
\pgftext[x=3.736547in,y=4.350693in,,bottom]{\color{textcolor}{\rmfamily\fontsize{9.000000}{10.800000}\selectfont\catcode`\^=\active\def^{\ifmmode\sp\else\^{}\fi}\catcode`\%=\active\def
\end{pgfscope}%
\begin{pgfscope}%
\definecolor{textcolor}{rgb}{0.000000,0.000000,0.000000}%
\pgfsetstrokecolor{textcolor}%
\pgfsetfillcolor{textcolor}%
\pgftext[x=1.262209in,y=3.462440in,,bottom]{\color{textcolor}{\rmfamily\fontsize{9.000000}{10.800000}\selectfont\catcode`\^=\active\def^{\ifmmode\sp\else\^{}\fi}\catcode`\%=\active\def
\end{pgfscope}%
\begin{pgfscope}%
\definecolor{textcolor}{rgb}{0.000000,0.000000,0.000000}%
\pgfsetstrokecolor{textcolor}%
\pgfsetfillcolor{textcolor}%
\pgftext[x=3.736547in,y=3.462440in,,bottom]{\color{textcolor}{\rmfamily\fontsize{9.000000}{10.800000}\selectfont\catcode`\^=\active\def^{\ifmmode\sp\else\^{}\fi}\catcode`\%=\active\def
\end{pgfscope}%
\begin{pgfscope}%
\definecolor{textcolor}{rgb}{0.000000,0.000000,0.000000}%
\pgfsetstrokecolor{textcolor}%
\pgfsetfillcolor{textcolor}%
\pgftext[x=1.262209in,y=2.574186in,,bottom]{\color{textcolor}{\rmfamily\fontsize{9.000000}{10.800000}\selectfont\catcode`\^=\active\def^{\ifmmode\sp\else\^{}\fi}\catcode`\%=\active\def
\end{pgfscope}%
\begin{pgfscope}%
\definecolor{textcolor}{rgb}{0.000000,0.000000,0.000000}%
\pgfsetstrokecolor{textcolor}%
\pgfsetfillcolor{textcolor}%
\pgftext[x=3.736547in,y=2.574186in,,bottom]{\color{textcolor}{\rmfamily\fontsize{9.000000}{10.800000}\selectfont\catcode`\^=\active\def^{\ifmmode\sp\else\^{}\fi}\catcode`\%=\active\def
\end{pgfscope}%
\begin{pgfscope}%
\definecolor{textcolor}{rgb}{0.000000,0.000000,0.000000}%
\pgfsetstrokecolor{textcolor}%
\pgfsetfillcolor{textcolor}%
\pgftext[x=1.262209in,y=1.685933in,,bottom]{\color{textcolor}{\rmfamily\fontsize{9.000000}{10.800000}\selectfont\catcode`\^=\active\def^{\ifmmode\sp\else\^{}\fi}\catcode`\%=\active\def
\end{pgfscope}%
\begin{pgfscope}%
\definecolor{textcolor}{rgb}{0.000000,0.000000,0.000000}%
\pgfsetstrokecolor{textcolor}%
\pgfsetfillcolor{textcolor}%
\pgftext[x=3.736547in,y=1.685933in,,bottom]{\color{textcolor}{\rmfamily\fontsize{9.000000}{10.800000}\selectfont\catcode`\^=\active\def^{\ifmmode\sp\else\^{}\fi}\catcode`\%=\active\def
\end{pgfscope}%
\begin{pgfscope}%
\definecolor{textcolor}{rgb}{0.000000,0.000000,0.000000}%
\pgfsetstrokecolor{textcolor}%
\pgfsetfillcolor{textcolor}%
\pgftext[x=1.262209in,y=0.797679in,,bottom]{\color{textcolor}{\rmfamily\fontsize{9.000000}{10.800000}\selectfont\catcode`\^=\active\def^{\ifmmode\sp\else\^{}\fi}\catcode`\%=\active\def
\end{pgfscope}%
\begin{pgfscope}%
\definecolor{textcolor}{rgb}{0.000000,0.000000,0.000000}%
\pgfsetstrokecolor{textcolor}%
\pgfsetfillcolor{textcolor}%
\pgftext[x=3.736547in,y=0.797679in,,bottom]{\color{textcolor}{\rmfamily\fontsize{9.000000}{10.800000}\selectfont\catcode`\^=\active\def^{\ifmmode\sp\else\^{}\fi}\catcode`\%=\active\def
\end{pgfscope}%
\end{pgfpicture}%
\makeatother%
\endgroup%